\documentclass[lettersize,journal]{IEEEtran}
\ifCLASSOPTIONcompsoc
  \usepackage[nocompress]{cite}
\else
  \usepackage{cite}
\fi
\ifCLASSINFOpdf
\else
\fi

\usepackage{bm}
\usepackage{amssymb}
\usepackage{latexsym}
\usepackage{dsfont}
\usepackage{multirow}
\usepackage{amsfonts}
\usepackage{amsmath}
\usepackage{algorithm}
\usepackage{algorithmic}
\usepackage{graphicx}
\usepackage{hyperref}
\usepackage{color}
\usepackage{soul} 
\usepackage{centernot}
\usepackage{mathtools}
\usepackage{ stmaryrd }
\usepackage{subfigure}

\DeclareMathAlphabet{\mathsfit}{\encodingdefault}{\sfdefault}{m}{sl}
\SetMathAlphabet{\mathsfit}{bold}{\encodingdefault}{\sfdefault}{bx}{n}

\DeclareMathOperator*{\argmax}{arg\,max}

\def\eg{{\it e.g.}}

\begin{document}
%
\title{A Survey on Text-guided 3D Visual Grounding: Elements, Recent Advances, and Future Directions}

\author{Daizong~Liu, Yang~Liu, Wencan~Huang, 
        and~Wei~Hu,~\IEEEmembership{Senior~Member,~IEEE}
\IEEEcompsocitemizethanks{\IEEEcompsocthanksitem D. Liu, Y. Liu, W. Huang and W. Hu are with Wangxuan Institute of Computer Technology, Peking University, No. 128, Zhongguancun North Street, Beijing, China. E-mail: dzliu@stu.pku.edu.cn, 2018liuyang@pku.edu.cn, huangwencan@stu.pku.edu.cn, forhuwei@pku.edu.cn. 
\IEEEcompsocthanksitem D. Liu and Y. Liu are Co-First authors. Corresponding author: Wei Hu. 
}}

%
%

\markboth{Journal of \LaTeX\ Class Files,~Vol.~14, No.~8, August~2015}%
{Shell \MakeLowercase{\textit{et al.}}: Bare Demo of IEEEtran.cls for Computer Society Journals}
%



\IEEEtitleabstractindextext{%
\begin{abstract}
Text-guided 3D visual grounding (T-3DVG), which aims to locate a specific object that semantically corresponds to a language query from a complicated 3D scene, has drawn increasing attention in the 3D research community over the past few years. Compared to 2D visual grounding, this task presents great potential and challenges due to its closer proximity to the real world, the complexity of data collection and 3D point cloud source processing.
In this survey, we attempt to provide a comprehensive overview of the T-3DVG progress, including its fundamental elements, recent research advances, and future research directions.
To the best of our knowledge, this is the first systematic survey on the T-3DVG task.
Specifically, we first provide a general structure of the T-3DVG pipeline with detailed components in a tutorial style, presenting a complete background overview. Then, we summarize the existing T-3DVG approaches into different categories and analyze their strengths and weaknesses. We also present the benchmark datasets and evaluation metrics to assess their performances. Finally, we discuss the potential limitations of existing T-3DVG and share some insights on several promising research directions.

\end{abstract}

}

\maketitle

\IEEEdisplaynontitleabstractindextext

%
\IEEEpeerreviewmaketitle

\section{Introduction}
\IEEEPARstart{T}{hree}-dimensional (3D) visual data has emerged as a pivotal component in daily life, thanks to advancements in sensing technologies and computer vision \cite{guo2020deep,bello2020deep,achlioptas2018learning}.
Unlike traditional 2D images, 3D scenes are generally represented as more complicated unordered point clouds, capturing richer spatial and depth information and introducing unique challenges and complexities due to the increased dimensionality and the intricacies of geometric and semantic interpretations.
Therefore, 3D scenes have natural advantages for multimedia intelligence exploration and research, drawing significant attention to related real-world applications such as robotic navigation \cite{wang2020heterogeneous,zhang2020continuous}, computer-aided room design \cite{sipe2002feature,ganin2021computer}, and human-computer interaction \cite{aggarwal2004human,li2020detailed,yu2010tree}. 
However, raw 3D scenes are often too redundant with multiple different types of objects, and of high information sparsity against the user-specific retrieval demands for finding one object. 
Furthermore, it is also challenging to maintain and manage these raw 3D scenes since they need to occupy a huge number of storage resources. 
Therefore, the ability to quickly retrieve a specific object from a 3D scene can allow users to locate highlighted objects of their interests conveniently and help information providers to optimize the storage fundamentally, thus being of great importance and interest in the research community.

\subsection{Definition and History}
\begin{figure}[t!]
\centering
\includegraphics[width=0.9\columnwidth]{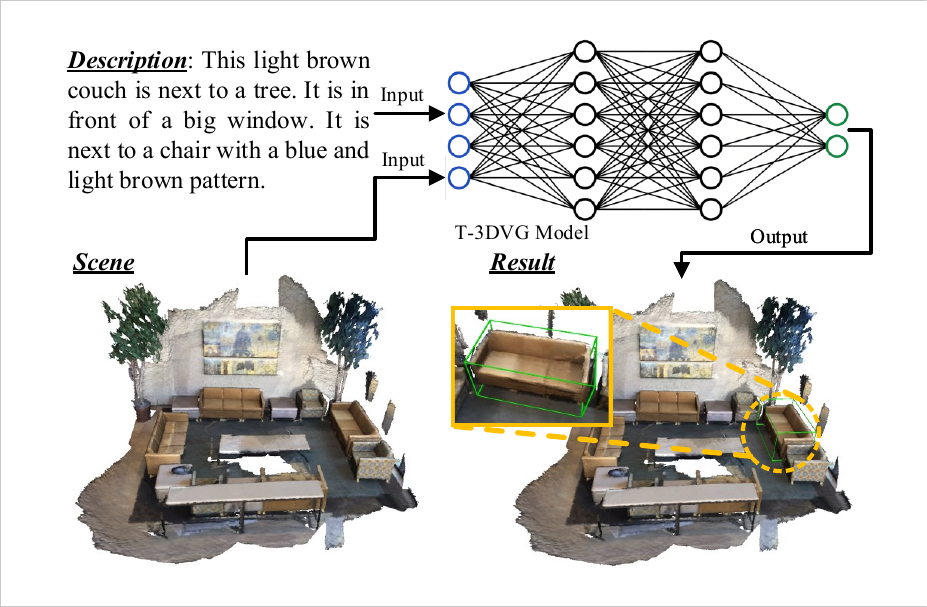}
\caption{An illustration of the text-guided 3D visual grounding (T-3DVG).}
    \label{fig:intro}
\end{figure}

Given a point-cloud-based 3D scene, the text-guided 3D visual grounding (T-3DVG) task \cite{chen2020scanrefer,achlioptas2020referit3d} aims to locate a specific 3D object semantically according to a free-form language description. For example, as shown in Fig.~\ref{fig:intro}, for the sentence description \textit{“This light brown couch is next to a tree. It is in front of a big window. It is next to a chair with a blue and light brown pattern”}, T-3DVG is required to predict the precise bounding box (the green one in the figure) of the corresponding object “couch”.

As a fundamental yet challenging vision-language task, T-3DVG serves as an intermediate step for various downstream 3D multimedia tasks, such as multi-modal 3D scene understanding \cite{lei2023recent,yu2023comprehensive}, language-guided robotic grasping \cite{tziafas2023language}, and 3D embodied interaction \cite{zhen20243d}. 
These tasks need to first localize relevant objects according to given queries (image or text), and then correlate the retrieved objects with the spatial-related backgrounds of the scene to reason and output suitable intermediate features, answers, or actions.
Therefore, T-3DVG plays a critical role in providing robust reasoning abilities for many existing 3D-related multi-modal tasks.

T-3DVG also shares similarities with some classic multi-modal tasks. For instance, text-guided 2D video grounding \cite{zhang2023temporal,lan2023survey,liu2023survey} and image grounding \cite{qiao2020referring,ji2024survey} aim to locate an activity-specific moment/object related to a sentence query from a video/image. These 2D tasks require matching the semantic features of video frames or images with the query to determine the moment boundary or object location, respectively.
In contrast, T-3DVG solely has coordinate-wise points instead of prior knowledge of any instance to match the sentence.
Besides, T-3DVG is also more challenging as scene data contains both complex spatial relationships and complicated background contents compared to 2D images. Therefore, T-3DVG needs to comprehensively understand the spatial relations of different objects from the textual description and reason them in scene data. T-3DVG is also similar to image-guided 3D detection \cite{qi2020imvotenet,li2008nonthreshold,jiang2022hypergraph,vora2020pointpainting,huang2020epnet}, which aims to locate a specific object in outdoor scenes by querying its 2D pictures. Since the same information on object appearance and shape is shared between images and point clouds, this task only needs a simple matching between cross-modal semantics. In comparison, T-3DVG is more challenging as its language guidance solely provides textual information but with complex spatial correlations, making it difficult to retrieve objects in complicated scenes due to the huge modality gap between vision and language.

T-3DVG was proposed in the year 2020 \cite{chen2020scanrefer,achlioptas2020referit3d}. The research works on this task published in the past few years can be roughly categorized into two types: two-stage and one-stage frameworks. Specifically, the two-stage framework \cite{roh2022languagerefer,chen2022language,he2021transrefer3d} first defines multiple object proposals through a pre-trained 3D detector, then reasons the detected objects with the language text to determine the best one. Although this type of framework has achieved significant performance, it relies on the quality of the 3D detectors and requires time-consuming reasoning among the multiple proposals while the frameworks are not end-to-end trained. Without using object proposals, the one-stage framework \cite{huang2023dense,luo20223d,wu2023eda} takes the whole 3D scene as the input and embeds point-level features in an end-to-end manner. After point-text cross-modal reasoning, it directly regresses the spatial bounding box of the text-guided object. However, this framework overlooks the rich information between the global points and sometimes fails to model the object-level correlations, thus their predicted object proposals are relatively coarse.
To address the above challenges, intensive works \cite{zhao20213dvg,abdelreheem20223dreftransformer} have been proposed to design better feature encoders and cross-modal interactors in the early stage. Later, additional contextual semantics (like 2D assistant data, multi-view data) \cite{yang2021sat,huang2022multi} are investigated to improve the robustness and generalization of the text-scene understanding. Different settings like fully- and weakly-supervised learning \cite{chen2020scanrefer,wang2023distilling} have also been explored by researchers for implementing this task in real-world applications.
Multiple types of architectures, such as CNN-based \cite{chen2020scanrefer}, MLP-based \cite{achlioptas2020referit3d}, and Transformer-based \cite{zhao20213dvg}, are further introduced to tackle the task from different perspectives.

\subsection{Motivation and Significance}

\begin{figure}[t!]
    \centering
    \subfigure{ \label{fig:year} 
    \includegraphics[width=0.46\columnwidth]{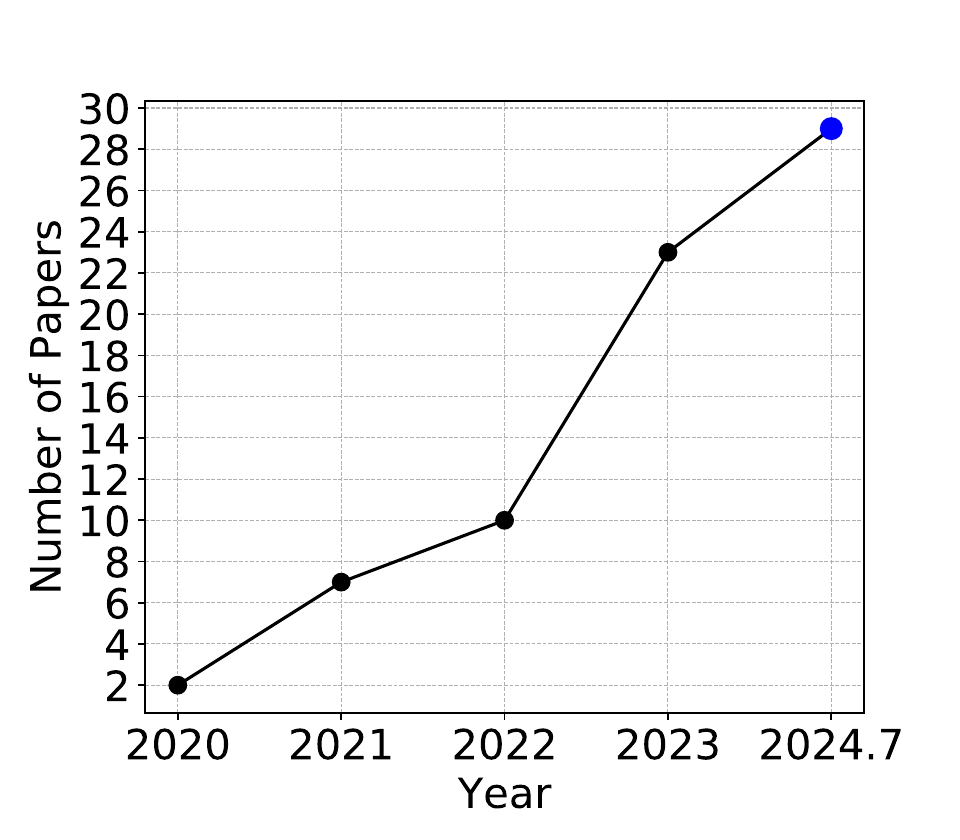}} 
    \hspace{-4pt}
    \subfigure { \label{fig:pub} 
    \includegraphics[width=0.50\columnwidth]{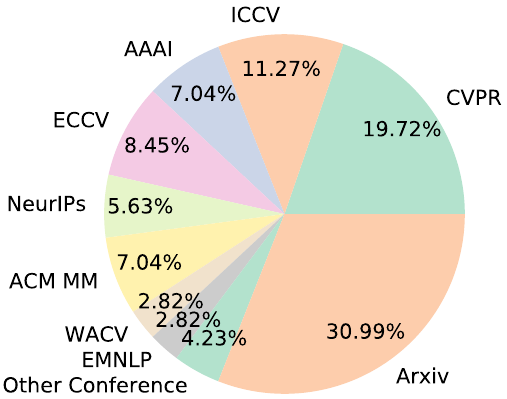}} 
    \caption{Statistics of the collected papers in this survey. Left: number of papers published in each year. Right: distribution of papers by venue.}
    \vspace{-4pt}
    \label{fig:statistic}
\end{figure}

However, without detailed guidance, the availability of a large number of existing T-3DVG methods can easily confuse researchers or practitioners trying to select or design an algorithm suitable for the specific problem at hand. To this end, this survey paper presents a comprehensive taxonomy of the recent approaches and developments of the text-guided 3D visual grounding research. 
We collect papers from reputable conferences and journals in Computer Vision, Natural Language Processing, Multimedia, and Machine Learning areas, \textit{e.g.}, CVPR, ECCV, ICCV, WACV, EMNLP, ACM MM, NeurIPS, AAAI, \textit{etc.}. The papers were mainly published from 2020 to 2024. Fig.~\ref{fig:statistic} summarizes the statistics of the collected papers.
By abstracting common generalities in all methods, we summarize the different types of 3D grounding methodologies and reveal a common pipeline of the T-3DVG model. We also give a detailed overview of the general elements in existing T-3DVG models and discuss their strengths and weaknesses.
At last, we conclude several concrete and promising future research directions (such as using powerful large models, and designing more practical settings) in this survey paper.

To be specific, we have three-fold motivations for this survey work:
\begin{itemize}
    \item Text-guided 3D visual grounding is an important task of multimedia understanding, which requires indicating a spatial-sensitive object in a complicated 3D scene via unrestricted natural language sentences. To the best of our knowledge, this is the first survey that constructs a taxonomy of 3D grounding techniques and elaborates methods in different categories with their strengths and weaknesses.
    \item It is of great interest to conduct in-depth discussion for important questions of text-guided 3D visual grounding: (1) How should the general 3D grounding data be processed? (2) What does a 3D grounding method generally look like and how it works? (3) What is the most intrinsic improvement of existing frameworks? (4) What are the problems that need to be solved in the future for this task?
    \item Comprehensively comparing and analyzing the experimental results on publicly available 3D grounding benchmarks would help readers to better understand the performance of each type of method as well as the corresponding network architecture.
\end{itemize}
Our goal is to provide the above perspectives on the composition of T-3DVG methods and the recent state of T-3DVG research. With such perspectives, researchers can not only confidently evaluate the trade-offs of various T-3DVG approaches in-depth and make informed decisions about using a suite of techniques to design a T-3DVG solution, but also appropriately implement suitable 3D grounding architecture in real-world applications.

The survey is organized as follows. In Section II, we introduce the general pipeline of existing T-3DVG methods and illustrate the technical details in a tutorial manner. It provides the reader with background knowledge about the functional components of T-3DVG models. Section III divides T-3DVG solutions into several categories, elaborates on the methods within each category, and discusses their advantages and disadvantages. 
Section IV summarizes the basic benchmark datasets and evaluation metrics. 
Section V summarizes the current research progress through performance comparison. Section VI discusses open issues and further research directions. Section VII concludes the paper.
\section{Background}
\begin{figure}[t!]
\centering
\includegraphics[width=1.0\columnwidth]{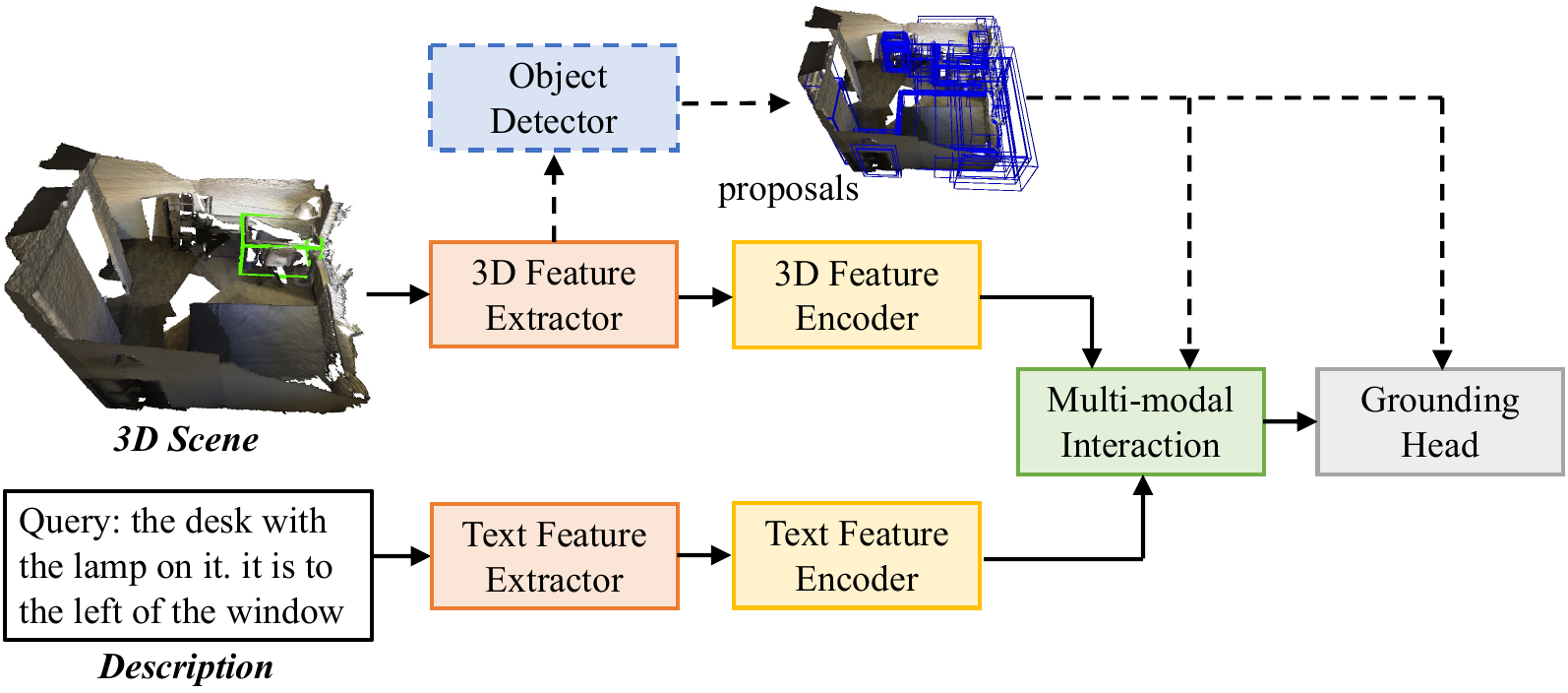}
\caption{A general pipeline for the T-3DVG task, consisting of multi-modal feature extractors, multi-modal feature encoders, multi-modal interaction module, and the final grounding head. An additional object detector is further introduced for two-stage T-3DVG methods to pre-extract all possible object proposals in each scene.}
    \label{fig:general_pip}
\end{figure}

Currently, there is no universal framework or process for text-guided 3D visual grounding (T-3DVG), and various methods propose novel architectures and techniques to address this problem from different perspectives. However, we observe that they often share common modules, as shown in Fig. \ref{fig:general_pip}. Before introducing the existing methods of T-3DVG, this section provides the general pipeline and details of some common tools (\textit{i.e.}, feature extractor, feature encoder, multi-modal interaction module, grounding head) as preliminaries.

\subsection{Basic Notations and Preliminary}
Before introducing each general module, we first define the basic notations involved in T-3DVG. 
T-3DVG typically takes inputs from two modalities. One is the scene point cloud $\mathcal{P}\in \mathbb{R}^{N\times (3+K)}$, where $N$ is the number of the points, and the feature dimension $3+K$ denotes three-dimensional coordinates and $K$-dimensional additional features (such as color, normals, height and multiview features). The other input is the descriptive text $\mathcal{T}=\{w_n\}_{n=1}^{L}$, where $w_n$ is the $n$-th word and $L$ is the number of the words. The task of T-3DVG aims to locate a specific object related to the semantic of $\mathcal{T}$ in the scene $\mathcal{P}$. The ground truth bounding box of the target object is denoted by $\mathcal{D} \in \mathbb{R}^{b}$, where the dimensions $b$ of $\mathcal{D}$ include a 6-dimensional 3D center point coordinates, 3D bounding box size, and an 18-dimensional one-hot label for the category (in some cases). Therefore, the goal of the T-3DVG can be formulated as: $\mathcal{F}_{T-3DVG}: (\mathcal{P},\mathcal{T}) \xmapsto{} \mathcal{D}$.

\subsection{3D Scene Feature Extractor}

Before being fed into the 3D model, the 3D scene is generally preprocessed by users for standardization, as different scenes often contain different numbers of points.
After that, existing T-3DVG methods either encode the semantics of 3D scenes via object detection models for object-level feature extraction, or directly encode the whole point-level 3D scene via the pre-trained 3D representation learning models. 

\noindent \textbf{Scene preprocessor.}
Generally, the number of points contained within the 3D scene point cloud (\eg, the commonly adopted ScanNet dataset \cite{dai2017scannet}) can be extremely large and variable. Processing each point in the original point cloud is typically costly and not conducive to parallel computation.
For example, each scene in the ScanNet dataset contains about 40 million points.
Therefore, it is reasonable to downsample all of them to a certain number of points for efficient processing.
Existing T-3DVG methods typically begin by randomly sampling $N_p$ points from the scene point cloud $\mathcal{P}$ as input points $\mathcal{P^\prime}\in \mathbb{R}^{N_p\times (3+K)}$  ($N_p$ is commonly set to 40,000 in practice). Then, the point cloud undergoes traditional data augmentation strategies consisting of random flipping, random rotation, random translation, and random size scaling for robust learning. 

\noindent \textbf{Object-level feature extractor.}
After preprocessing the original scene data, existing T-3DVG methods with a two-stage framework generally utilize a pre-trained 3D object detector or segmentor to extract candidate objects and encode them at the object level. 
These object-level features are utilized to interact with the sentence query for reasoning the best matched object.
Commonly used 3D object detectors include Votenet \cite{qi2019deep}, GroupFree \cite{liu2021group}, PointGroup \cite{jiang2020pointgroup}, \textit{etc.}.
Some other methods propose extracting instances from the point cloud using a segmentation model \cite{kirillov2019panoptic,han2020occuseg,engelmann2017exploring,engelmann2018know,xie2020linking} to obtain more fine-grained features.
Specifically, these two-stage T-3DVG methods typically detect/segment all possible candidate objects in the point cloud and obtain their localization $\mathcal{D} \in \mathbb{R}^{M\times 6}$ and their features $\mathbf{C}_i \in \mathbb{R}^{M\times d}$, where $M$ is the number of candidate objects and $d$ is the dimension of the objects' feature. 
The localization of each object $\mathcal{D}_i \in \mathbb{R}^{6}=(c_x,c_y,c_z,r_x,r_y,r_z)$ is represented by its 3D center coordinates $\mathbf{c}$ and 3D box length $\mathbf{r}$. 
Overall, the feature extraction process can be described as:
\begin{equation}
\begin{aligned}
    &\mathcal{P^\prime}\in \mathbb{R}^{N_p\times (3+K)} \xmapsto[\text{detector}]{\text{encoder}} \{\mathcal{D}_i \in \mathbb{R}^6 \}_{i=1}^{M} \\
    &\xmapsto[\text{extractor}]{\text{object feature} } \{\mathbf{C}_i \in \mathbb{R}^d \}_{i=1}^{M}.
\end{aligned}
\end{equation}


\noindent \textbf{Point-level feature extractor.}
Existing T-3DVG methods with a single-stage framework does not explicitly detect or segment objects.
Instead, these methods directly utilize a global point cloud encoder, usually employing PointNet++ \cite{qi2017pointnet++} or 3D convolutions \cite{choy20194d}, to obtain point-level features, which are then used as extracted scene features. 
Compared to the object-level features, these point-level features are more fine-grained and can perceive more relevant background contexts. However, it also introduces much irrelevant background noise for ineffective reasoning.
The point-level feature extraction process can be described as:
\begin{equation}
    \mathcal{P^\prime}\in \mathbb{R}^{N_p\times (3+K)} \xmapsto{\text{encoder}} \mathbf{C} \in \mathbb{R}^{M\times d},
\end{equation}
where $M$ is the number of the seed points and $d$ is the dimension of the point-level features.

\subsection{Text Feature Extractor}
\noindent \textbf{Text preprocessor.}
The textual inputs $\mathcal{T}=\{w_n\}_{n=1}^{L_w}$ provided in the dataset consist of sentences of varying lengths.
For preprocessing, each text query is typically segmented into tokens. Sentences longer than a certain threshold are truncated, and those shorter than the threshold are padded with zeros. Generally, existing T-3DVG methods set this threshold to the same number of $L$. Subsequently, existing methods typically employ textual data augmentation by randomly replacing some tokens corresponding to object categories in sentences with specific \textit{Unknown token} [UNK]. Some methods employ additional data augmentation techniques, such as randomly masking out words \cite{zhao20213dvg} or combining multiple sentences within the same scene \cite{chen2022ham}. Some other methods \cite{guo2023viewrefer} utilize large language models (LLM) to expand input text, thereby providing supplementary information, such as generating textual descriptions from different viewpoints compared to the original semantics.

\noindent \textbf{Text feature extractor.}\label{sec:text_extractor}
After preprocessing the textual input, the query is extracted with a feature extractor to map its semantics into an embedding space. Common embedding methods used in T-3DVG include pre-trained word embeddings, such as GloVe \cite{pennington2014glove}, and pre-trained language models, such as RoBERTa \cite{liu2019roberta}. The whole feature extracting process can be formulated as:
\begin{equation}
\begin{aligned}
    &\mathcal{T}=\{w_n\}_{n=1}^{L_w} \xmapsto[\text{or pad}]{\text{truncate}} \mathcal{T}^\prime=\{w_n\}_{n=1}^{L} \\
    &\xmapsto[\text{extractor}]{\text{text feature} } \mathbf{T} \in \mathbb{R}^{L\times d_t},
\end{aligned}
\end{equation}
where $d_t$ is the dimension of text features.

Some T-3DVG methods \cite{wu2023eda,feng2021free} further utilize off-the-shelf natural language processing tools \cite{schuster2015generating,wu2019unified} to parse the text, decomposing it into target objects, auxiliary object, object attributes, relationships, \textit{etc.}. This process facilitates finer utilization of textual information.

\subsection{Scene and Text Context Encoder}
After separately extracting rough features from the scene point cloud and text, many T-3DVG methods further design intra-modal encoders to capture the dependencies within each modality to obtain richer contextual information. Based on these more contextual features, in the latter process, they then map information from different modalities into the same space to facilitate subsequent interaction and matching. Despite using these contextual encoders, some early T-3DVG methods omit this encoding step and directly fuse the extracted rough features of the point cloud and text for subsequent processing.

\noindent \textbf{Scene context encoder.}
\begin{figure}[t!]
\centering
\includegraphics[width=0.9\columnwidth]{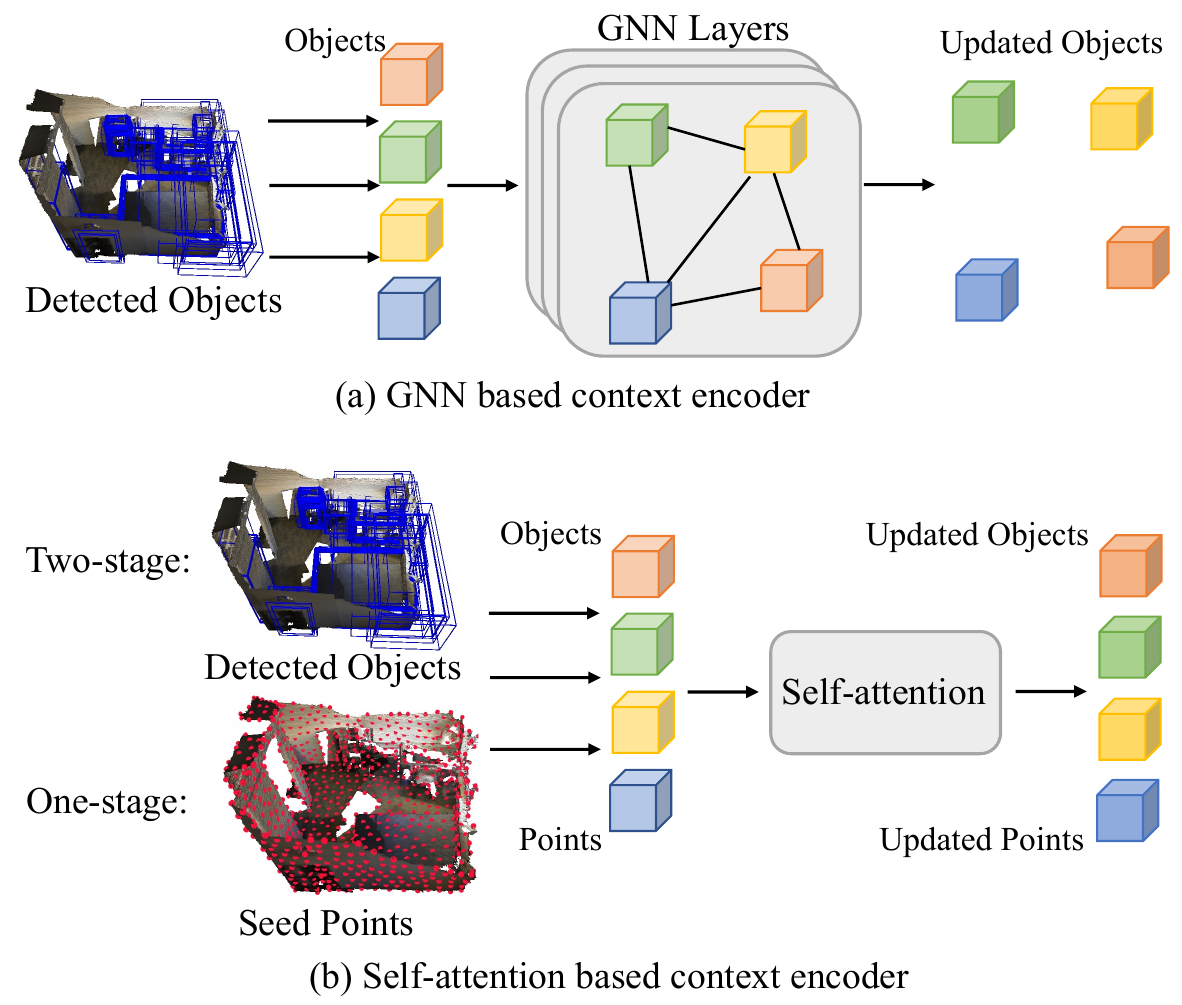}
\caption{Illustration of two types of scene context encoders. They learn more dependencies among intra-modal instances (objects or points).}
    \label{fig:bg1}
\end{figure} 
The common scene context encoders can be categorized into two main types: the approaches \cite{huang2021text,feng2021free} based on graph neural networks \cite{kipf2016semi} and the approaches \cite{he2021transrefer3d,zhao20213dvg} based on self-attention mechanisms \cite{vaswani2017attention}, as illustrated in Fig. \ref{fig:bg1}. 
Scene encoders based on graph neural networks are mostly employed in the two-stage T-3DVG framework. It explicitly establishes a graph structure over the object proposals obtained through 3D detectors. Each node on the graph represents an object proposal and each edge represents the relationship between the proposals. Typically, this graph is constructed based on the spatial and semantic distances between proposals, with edges connecting nearby proposals. After constructing the proposal graph, Graph Convolutional Networks \cite{kipf2016semi} or other similar graph neural networks \cite{veličković2018graph,NIPS2017_5dd9db5e} are employed to encode the relationships between objects into the features of the proposals.
Besides, scene encoders based on self-attention mechanisms represent another popular approach that can handle both object-level and point-level visual features. Unlike graph-neural-network-based encoders, self-attention mechanisms operate without an explicit graph structure, allowing for automatic semantic correlation learning among object proposals. Some scene encoders \cite{luo20223d} based on self-attention mechanisms also incorporate attention between object features and point features to further refine object features from the points contained within them.
Some T-3DVG methods \cite{cai20223djcg,zhu20233d} also combine the advantages of these two designs, which manually inject spatial relationships between objects into self-attention mechanisms through methods such as position encoding or attention modulation. Additionally, some methods \cite{huang2022multi,bakr2022look,hsu2023ns3d} propose to enrich proposal information by adding multi-view information or encoding ternary relationships. Overall, the general scene context encoding process can be described as:
\begin{equation}
    \{\mathbf{C}_i \in \mathbb{R}^d \}_{i=1}^{M} \ \ or \ \ \mathbf{C} \in \mathbb{R}^{M\times d} \mapsto
    \mathbf{C^\prime} \in \mathbb{R}^{M\times d_c},
\end{equation}
where $d_c$ is the dimension of contextual scene features generated by the scene encoder.

\noindent \textbf{Text context encoder.}
Common text context encoders can be categorized into two types according to existing T-3DVG methods: the approaches \cite{achlioptas2020referit3d,chen2020scanrefer,huang2021text} based on recurrent neural networks and the approaches \cite{he2021transrefer3d,zhao20213dvg,abdelreheem20223dreftransformer} utilizing self-attention mechanisms. As for the first type of approaches, their popular recurrent neural network encoders include GRU \cite{chung2014empirical} and LSTM \cite{hochreiter1997long}, which effectively capture temporal dependencies in the text. 
These recurrent neural network encoders can further be classified into two categories based on their textual output: word-level encoders and sentence-level encoders. Word-level encoders maintain the dimensionality corresponding to the number of words in their output, while sentence-level encoders encode an entire sentence into a single vector feature as output. The following illustrates both kinds of encoders:
\begin{equation}
\begin{aligned}
    &\mathbf{T} \in \mathbb{R}^{L\times d_t}  \xmapsto[\text{RNN}]{\text{word-level}}
    \mathbf{T^\prime} \in \mathbb{R}^{L\times d_w},\\
    &\mathbf{T} \in \mathbb{R}^{L\times d_t}  \xmapsto[\text{RNN}]{\text{sentence-level}}
    \mathbf{T^\prime} \in \mathbb{R}^{d_s},
\end{aligned}
\end{equation}
where $d_w$ and $d_s$ represent the dimensions of word-level features and sentence-level features, respectively.

As for the second type of approaches, they utilize general self-attention \cite{vaswani2017attention} that is capable of capturing long-range dependencies in sentences more efficiently, albeit potentially requiring greater computational resources. There are some methods \cite{zhao20213dvg} that directly utilize a combination of these two types of encoders for contextual learning. 
Other methods also employ more powerful large-model encoders such as CLIP \cite{radford2021learning} for semantic learning. As described in Section~\ref{sec:text_extractor}, some methods also utilize pre-trained language models like RoBERTa \cite{liu2019roberta} as text feature extractors, which also includes encoding of textual context relationships. Overall, the encoder using self-attention or pre-trained model is typically at the word level, and its computational process can be represented as follows:
\begin{equation}
    \mathbf{T} \in \mathbb{R}^{L\times d_t}  \xmapsto[\text{or pre-trained model}]{\text{self-attention}}
    \mathbf{T^\prime} \in \mathbb{R}^{L\times d_w}.
\end{equation}

\subsection{Multi-Modal Interaction}
After separately encoding contextual scene and text features, the next step is to integrate and interact these two modal features for alignment and reasoning.
This requires fully comprehending the complete semantic meaning of the text query and the complex relations among multiple objects in the 3D scene, thus locating the best matched object that is most relevant to the query semantics based on the fused multimodal representation.
This is a critical step in T-3DVG, where many methods propose highly innovative model designs at this stage.

Some early T-3DVG methods \cite{achlioptas2020referit3d,chen2020scanrefer,liu2021refer,huang2021text,feng2021free} directly employ a simple concatenation operation to coarsely combine scene features and text features that have already been embedded in the same space. Considering such modality fusion based on concatenation cannot finely integrate and match the detailed features of the two modalities, some post-processing is usually required to merge the features of both modalities into the same feature space. Common post-processing methods include multilayer perceptron (MLP) \cite{chen2020scanrefer} and graph neural networks (GNN) \cite{achlioptas2020referit3d,huang2021text,feng2021free}, which are utilized to further refine the fused representations. There are also some methods that utilize attention modules or large models to achieve this. This type of multi-modal interaction module can be represented as follows:
\begin{equation}
    \{\mathbf{C^\prime} \in \mathbb{R}^{M\times d_c},\mathbf{T^\prime}\}  \xmapsto{\text{concat}}
    \mathbf{C}_{\text{cat}} \xmapsto[\text{process}]{\text{post-}} \mathbf{C}_{\text{fus}} \in \mathbb{R}^{M\times d_f},
\end{equation}
where $d_f$ is the dimension of fused features.

Besides, most existing methods \cite{he2021transrefer3d,zhao20213dvg,luo20223d,huang2022multi} employ a transformer-based cross-attention mechanism \cite{vaswani2017attention} as the core of cross-modal feature fusion. Within the cross-attention mechanism, object-level features or point-level features typically serve as the attentive query, while text features function as the attentive key and attentive value. This configuration allows for the assessment of the correspondence between each object/point and the query text. Some methods \cite{abdelreheem20223dreftransformer} also incorporate dual-pathway cross-encoder modules to additionally obtain the associations between each word-object pair, where text features serve as attentive queries and object or point features serve as attentive keys and attentive values. To further improve the attention performance, many methods have made modifications to the cross-attention mechanism, such as masking the attention scope to capture the most relevant object relations \cite{unal2023three} or performing cross-attention at both the sentence and word levels \cite{chen2022learning}. Some methods incorporate additional information to help cross-attention capture richer information. For example, 3DVG-Transformer \cite{zhao20213dvg} multiplies spatial features with the attention matrix, while 3D-SPS \cite{luo20223d} uses the initial point feature as another input to the cross-attention.

\begin{figure}[t!]
\centering
\includegraphics[width=0.9\columnwidth]{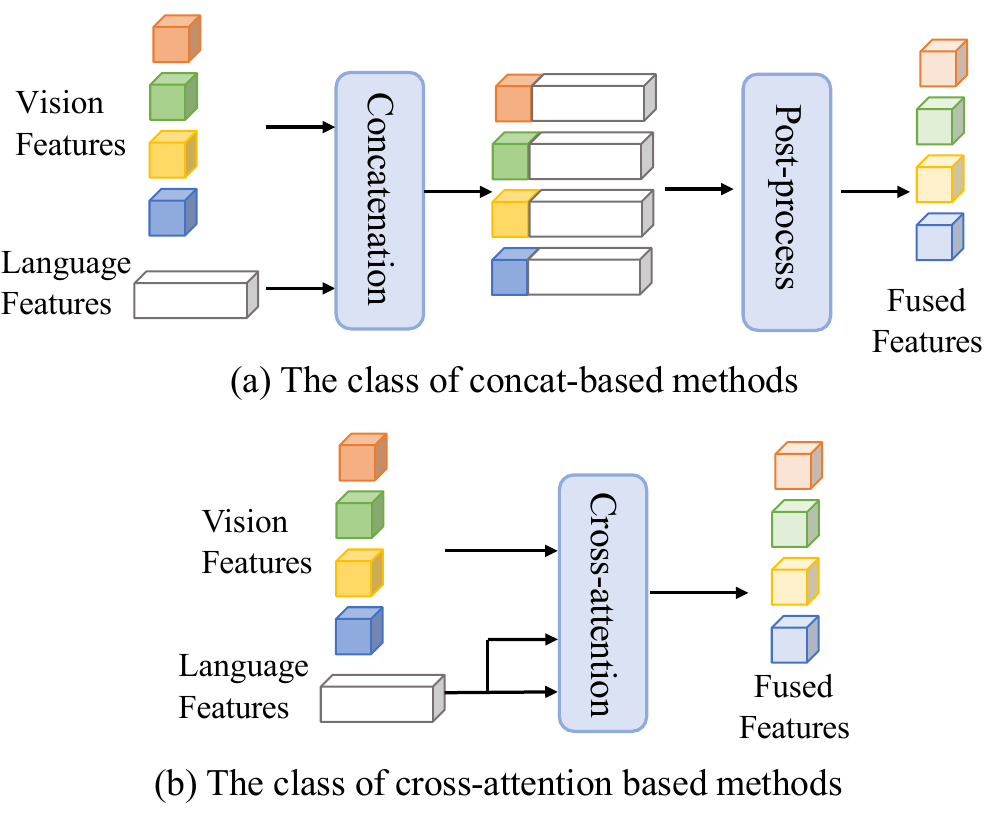}
\vspace{-14pt}
\caption{Illustration of two types of multi-modal interaction methods. }
\vspace{-10pt}
    \label{fig:bg3}
\end{figure} 

In addition to the multi-modal attentive fusion, the existing methods are also coupled with individually designed pre-processing modules.
The main purpose of this pre-processing is to generate better attentive queries, attentive keys, and attentive values in the attention mechanism. 
A common pre-processing approach involves an initial filtering of objects or points, selecting those closely related to the query text based on simple criteria. This filtering can be achieved through explicit filtering based on object categories \cite{yuan2021instancerefer} or through implicitly neural network-based learning \cite{luo20223d,chen2022learning}.
Some other methods \cite{jin2023context} employ multi-modal alignment mechanisms as pre-processing, which aligns information from the two modalities at different levels to enable the cross-attention mechanism to better fuse language and visual information. This types of multi-modal interaction module can be represented as follows:
\begin{equation}
    \{\mathbf{C^\prime} \in \mathbb{R}^{M\times d_c},\mathbf{T^\prime}\}  \xmapsto{\text{cross-attention}}
    \mathbf{C}_{\text{fus}} \in \mathbb{R}^{M\times d_f} .
\end{equation}

The two types of multi-modal interaction methods are illustrated in Fig. \ref{fig:bg3}.

\subsection{Grounding Head and Training Objectives}
\begin{figure}[t!]
\centering
\includegraphics[width=0.9\columnwidth]{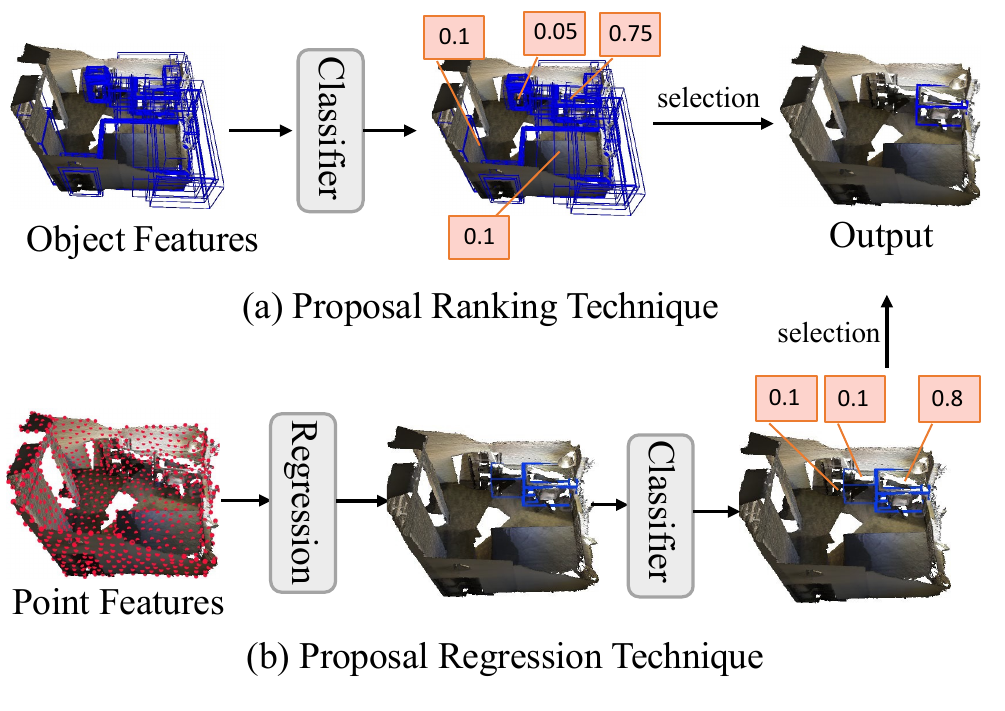}
\vspace{-10pt}
\caption{Illustration of two types of grounding heads. }
    \label{fig:bg4}
\end{figure}

The last stage of T-3DVG utilizes a grounding head to transform the features obtained from the cross-modal interaction stage into the final output. T-3DVG methods employing a two-stage framework typically utilize a Proposal Ranking Technique to accomplish the grounding task, whereas those using a one-stage framework tend to adopt a Proposal Regression Technique. The grounding heads of both approaches can often be realized with concise MLP layers. The two types of grounding head are shown in Fig. \ref{fig:bg4}.
\subsubsection{Proposal Ranking Technique}
For methods using a two-stage framework, since the location of all possible objects has been obtained through a detector or segmentor, it is only necessary to sort the matching degree between these objects and the text through a Proposal Ranking Technique. Given the object location $\mathcal{D} \in \mathbb{R}^{M\times 6}$ obtained by the detector or segmentor, the existing methods use a classifier to calculate a confidence score $\mathbf{s} \in \mathbb{R}^M$ for each object based on the fused features $\mathbf{C}_{\text{fus}}$, representing the matching degree between the object and the query text, that is, the possibility that a certain object is the target object. The classifier is usually implemented using an MLP with a sigmoid function. Finally, the location of the object $\mathcal{D}_j \in \mathbb{R}^6$ with the highest confidence score $j=\argmax(\mathbf{s})$ is output as the grounding result.

\subsubsection{Proposal Regression Technique}
For methods using a one-stage framework, as they do not have the location of the objects, this step requires employing a Proposal Regression Technique to regress the target object's position based on the multi-modal reasoning features $\mathbf{C}_{\text{fus}}$. In one way, this can be achieved by first regressing the positions of multiple objects along with their confidence scores, then selecting the one with the highest confidence score as the final output \cite{luo20223d,wang20233drp}. Algorithms such as Hungarian matching \cite{kamath2021mdetr,jain2022bottom} are usually employed to obtain the object with the highest matching degree.
There are also some works \cite{huang2023dense,yuan2022toward,wu20243d} that regress multiple objects and further reason their relations for final determining.

\subsubsection{Training Objectives}
The training objective of the T-3DVG task usually involves a combination of multiple losses. 

As for the two-stage technique,
the primary training objective is the object grounding loss $\mathcal{L}_{\text {VG}}$, which is usually implemented using the cross-entropy function to supervise the confidence score for each possible object:
\begin{equation}
    \mathcal{L}_{\text {VG}}=-\sum_{i=1}^M \mathbf{t}_i \log \left(\mathbf{s}_i\right).
\end{equation}
Here, $\mathbf{t}$ denotes the one-hot vector that represents the ground truth for all object proposals. The most common practice is to set $\mathbf{t}_j$ to 1 for the target object and others to 0 for background objects, where $j$ is the index of the proposal with the highest IoU with the ground truth. Some other methods employ a multi-hot vector as the ground truth, setting the indices with IoU values above a certain threshold to 1. Additionally, some methods \cite{yuan2021instancerefer} also adopt a contrastive learning strategy, treating instances with IoU above a certain threshold as positive examples $Q_i^{+}$ and others as negative examples $Q_i^{-}$ for learning more representative features and confidence scores as:
\begin{equation}
\mathcal{L}_{\text {VG}}=-\log \frac{\sum_{i=1}^L \exp \left(Q_i^{+}\right)}{\sum_{i=1}^L \exp \left(Q_i^{+}\right)+\sum_{i=L+1}^M \exp \left(Q_i^{-}\right)}.
\end{equation}

In addition to the primary training objective of object grounding, many methods \cite{achlioptas2020referit3d,chen2020scanrefer} also design some auxiliary tasks for robust representation learning, such as object detection, text classification, \textit{etc.}. The object detection loss is determined by the detector or segmentor used. For example, for methods that use VoteNet \cite{qi2019deep} as the detector, the object detection loss $\mathcal{L}_{\text {det}}$ is obtained by weighted summation of the vote regression loss $\mathcal{L}_{\text {vote}}$ \cite{qi2019deep}, objectness binary classification loss $\mathcal{L}_{\text {obj}}$, box regression loss $\mathcal{L}_{\text {box}}$, and the semantic classification loss $\mathcal{L}_{\text {sem-cls}}$, with balanced weights $\lambda$ as:
\begin{equation}
    \mathcal{L}_{\text {det}} = \mathcal{L}_{\text {vote}}+\lambda_1 \mathcal{L}_{\text {obj}}+\lambda_2 \mathcal{L}_{\text {box}}+\lambda_3 \mathcal{L}_{\text {sem-cls}}.
\end{equation}
Another commonly used auxiliary task is the language classification task, which trains a classifier to predict object categories from descriptive text to assist in training for text understanding. The language classification loss $\mathcal{L}_{\text {lang}}$ is typically implemented using cross-entropy.
Besides these two most common auxiliary tasks, many methods have designed some training objectives specifically for the new modules they propose, such as alignment loss to facilitate the alignment of textual and visual features \cite{wu2023eda,jin2023context}, and mask loss to train language models \cite{roh2022languagerefer}, etc.

As for the one-stage technique, similar losses in the two-stage technique are also employed to supervise feature learning. The main difference is that the one-stage technique only requires the confidence scoring loss and a regression loss $\mathcal{L}_{\text{regress}}$ \cite{luo20223d} to supervise the quality of the generated object.

\begin{figure*}[!tbp]
\centering
\includegraphics[width=2.0\columnwidth]{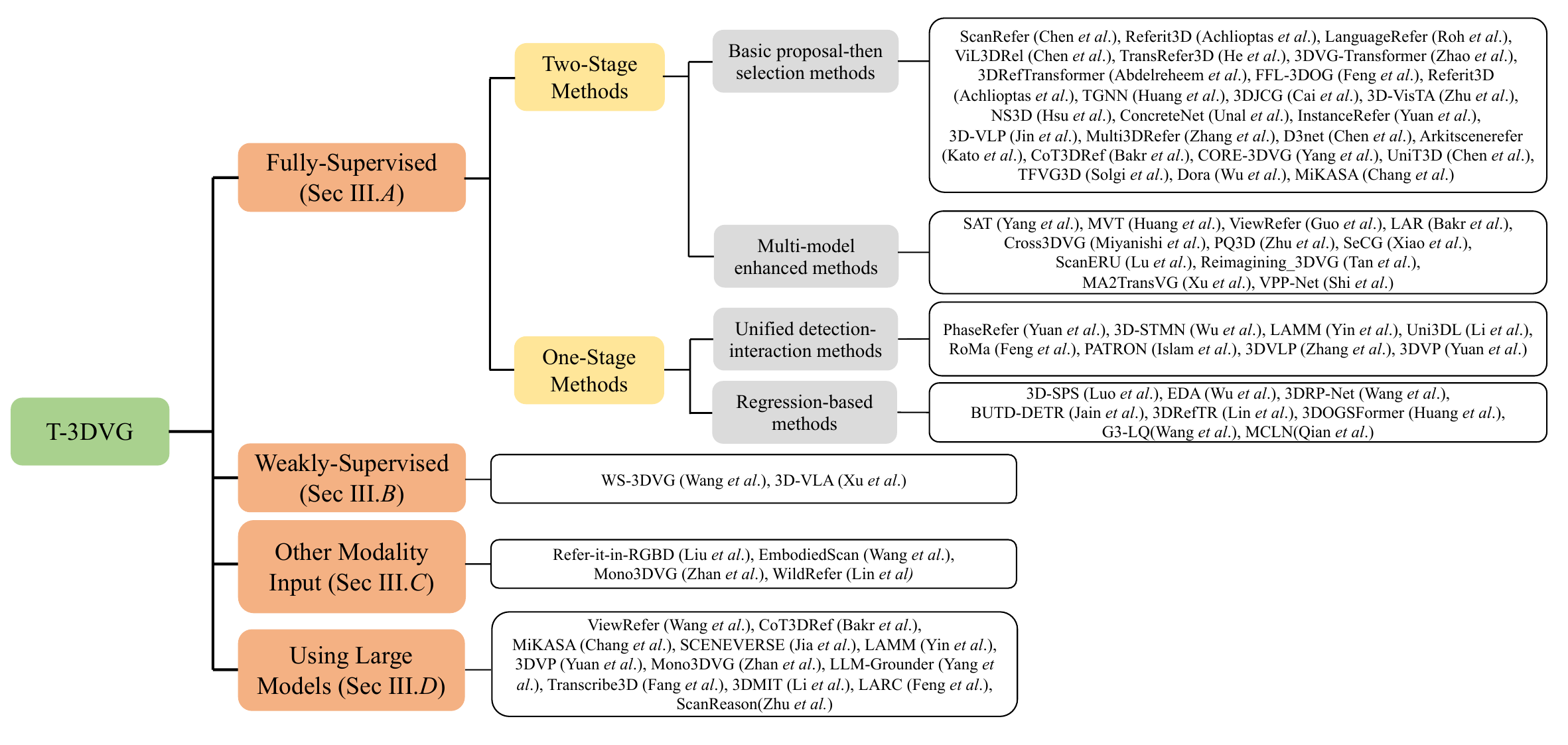}
\vspace{-10pt}
\caption{The taxonomy of T-3DVG based on different methods. The fully-supervised methods and weakly-supervised methods will be introduced in Sections III.\textit{A} and III.\textit{B}. The methods of using other modality input and large models will be described in Sections III.\textit{C} and III.\textit{D}. 
}
    \label{fig:method4}
\end{figure*} 

To summarize, in this section, we provide a brief review of the main components of the existing T-3DVG methods, including multi-modal feature extractors, multi-modal feature encoders, multi-modal interaction module, and the final grounding head. 
Although recent T-3DVG methods may contain more complex structures and different auxiliary modules, their model frameworks generally follow our summarized general T-3DVG pipeline.
\section{Methods}
Most of the existing T-3DVG methods are proposed with the fully-supervised learning paradigm. Early approaches primarily utilize pre-trained object detectors or segmentation models to pre-sample all potential objects from the complicated 3D scene. Then, these object proposals are then paired with a given sentence query to learn their semantic correspondence through cross-modal interaction to determine the most matching one. However, this two-stage framework not only severely relies on the quality of the 3D detectors and segmentation models, but also requires time-consuming reasoning among the multiple proposals. To alleviate this limitation, without using object proposals, the one-stage framework is proposed to take the whole 3D scene as the input and embed point-level features in an end-to-end manner. After point-text cross-modal reasoning, it directly regresses the spatial bounding box of the text-guided object.
Although the above two types of fully-supervised methods have achieved significant
performance in recent years, they require a large
number of bounding boxes for numerous objects in the 3D point
cloud scenes, along with their descriptive texts, to provide reliable supervision.
Considering manually annotating these object-text pairs is very time-consuming and labor-intensive, some works propose to solve the T-3DVG task in the weakly-supervised setting, which only utilizes text labels for grounding without relying on any object bounding box annotations.
There are also some recent works that utilize other-modal input (instead of point cloud) for grounding or incorporate traditional grounding models with powerful large language models.

In this section, we provide a complete review of existing T-3DVG methods via a taxonomy in Fig.~\ref{fig:method4}, based on the methods' architectures and learning algorithms, to better categorize T-3DVG approaches. Specifically, we discuss the unique characteristics of each category and illustrate the highlights of each T-3DVG method in details.

\subsection{Fully-Supervised Methods}
\subsubsection{Two-Stage Methods}
Most existing T-3DVG methods \cite{chen2020scanrefer,achlioptas2020referit3d,yuan2021instancerefer,huang2021text,zhao20213dvg,abdelreheem20223dreftransformer,feng2021free,roh2022languagerefer,chen2022language,he2021transrefer3d,cai20223djcg,chen2022d,zhu20233d,kato2023arkitscenerefer,jin2023context,bakr2023cot3dref,yang2024exploiting,hsu2023ns3d,zhang2023multi3drefer,unal2023three,chen2023unit3d,solgi2024transformer,wu2024dora,chang2024mikasa,yang2021sat,bakr2022look,miyanishi2023cross3dvg,zhu2024unifying,tan2023reimagining,guo2023viewrefer,xu2024multi,shi2024aware,xu2024dual} are deployed in a two-stage framework, which first utilizes a pre-trained 3D detection or segmentation model to obtain all possible 3D objects in the scene and then individually matches and reasons them with the query for selecting the best one.
These works can be further categorized into a basic proposal-then-selection approach and a multi-modal enhanced proposal-then-selection approach. In particular, the latter approach integrates more contextual modalities' information into the 3D reasoning process for enriching the representation learning.

\begin{figure}[t!]
\centering
\includegraphics[width=1.0\columnwidth]{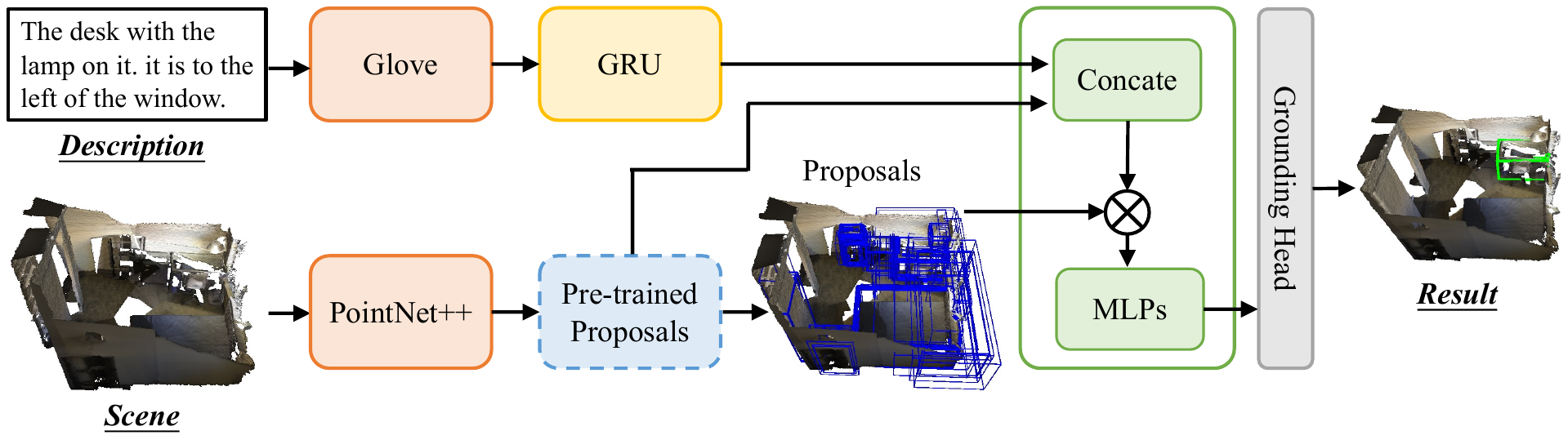}
\caption{ScanRefer architecture, reproduced from Chen \textit{et al.} \cite{chen2020scanrefer}.}
    \label{fig:method2}
\end{figure}

\noindent \textbf{Basic proposal-then-selection methods.}
ScanRefer \cite{chen2020scanrefer} (as shown in Fig.~\ref{fig:method2}) and ReferIt3D \cite{achlioptas2020referit3d} are the pioneer methods proposed to address the T-3DVG task following a basic proposal-then-selection methods. They generally first utilize a pre-trained 3D object detector to obtain all possible objects within the whole 3D scene. Then, they match the objects' features with the semantic of text descriptions for determining the best aligned object as the final output.
Instead of using pre-trained 3D objectors, a few works \cite{yuan2021instancerefer,huang2021text} also introduce to utilize a pre-trained segmentor to obtain fine-grained semantic features for text matching. 
Considering the object detector is relatively lightweight and enough to capture object-level features, almost all two-stage methods follow the detection-based framework.

Since the object proposals and features are fixed without additional need for training, 
a lot of works \cite{zhao20213dvg,abdelreheem20223dreftransformer,feng2021free,bakr2023cot3dref,yang2024exploiting,solgi2024transformer,chang2024mikasa} put more effort into designing effective intra- and inter-modal interactions.
Zhao \textit{et al.} \cite{zhao20213dvg}, Solgi \textit{et al.} \cite{solgi2024transformer} and Chang \textit{et al.} \cite{chang2024mikasa} propose to develop transformer-based architecture \cite{vaswani2017attention} to capture contextual object relations and achieve object-text reasoning instead of using simple attention mechanisms.
Abdelreheem \textit{et al.} \cite{abdelreheem20223dreftransformer} also propose an end-to-end transformation structure but with a spatial reasoning module to handle object-level reasoning.
Feng \textit{et al.} \cite{feng2021free} introduce the graph neural network \cite{wu2020comprehensive} into the T-3DVG framework to achieve the contextual self- and cross-modal interactions. As for the self-modal graph, object/word serves as nodes and their edge vector denotes the relations between adjacent notes. As for cross-modal graph, both object and word can serve as nodes for reasoning their matching scores.
To reason the object with the sentence in complex scenarios, Bakr \textit{et al.} \cite{bakr2023cot3dref} formulate the 3D visual grounding problem in a sequence-to-sequence fashion. It first predicts a chain of anchors to decompose the referring task into interpretable intermediate steps, and then generates the final target based on them, achieving fine-grained grounding performance. 
Yang \textit{et al.} \cite{yang2024exploiting} further disentangle the attention mechanism into a three-step reasoning process, \textit{i.e.}, text-guided object detection, relation matching network, and target identification, making the model more interpretable with better performance.
Chen \textit{et al.} \cite{chen2022ham} introduce
a novel hierarchical attention model to offer multi-granularity representation for better cross-modal reasoning. 

Besides, there are also some works \cite{roh2022languagerefer,chen2022language,he2021transrefer3d,kato2023arkitscenerefer,hsu2023ns3d,zhang2023multi3drefer,wu2024dora,liu2024cross,huang2024advancing} designed to address the task-specific challenges in T-3DVG.
Considering that object locations are important to reason the spatial relations mentioned in the sentence, spatial embeddings \cite{roh2022languagerefer,chen2022language} are designed to embed positional-encoded bounding box information into the attention mechanism.
To help distinguish complicated background objects from the foreground target, He \textit{et al.} \cite{he2021transrefer3d} propose to separately align low-level and high-level of words with the objects for comprehending and recognizing the corresponding object entity and relation.
Kato \textit{et al.} \cite{kato2023arkitscenerefer} introduce the ARKitSceneRefer method to tackle the object location on small indoor 3D objects. It introduces a new small-object dataset and develops a simple coarse-to-fine framework to achieve the goal.
Hsu \textit{et al.} \cite{hsu2023ns3d} propose to translate language into programs with hierarchical structures by leveraging large language-to-code models and extend prior neuro-symbolic visual reasoning methods by introducing functional modules that effectively reason about high-arity relations, key in disambiguating objects in complex 3D scenes. This framework can alleviate two issues: the expense of labeling and the complexity of 3D grounded language.
Zhang \textit{et al.} \cite{zhang2023multi3drefer} introduce a new case when multiple objects exist in the same scene, and devise a contrastive learning strategy to distinguish the objects for better determining.
Considering unstructured natural utterances and scattered objects might lead to undesirable performances, Wu \textit{et al.} \cite{wu2024dora} propose to leverage language models to parse language description, suggesting a referential order of anchor objects. Such ordered anchor objects allow the model to update visual features and locate the target object during the grounding process.
Instead of localizing a unique pre-existing object within a single 3D scene like previous works, Huang \textit{et al.} \cite{huang2024advancing} introduce a more realistic setting, named Group-wise 3D Object Grounding, to simultaneously process a group of related 3D scenes, allowing a flexible number of target objects to exist in each scene.

Moreover, some proposal-then-selection works \cite{cai20223djcg,chen2022d,zhu20233d,jin2023context,unal2023three,chen2023unit3d} design novel training strategies or relevant joint tasks to assist 3D grounding.
For example,
Cai \textit{et al.} \cite{cai20223djcg} and Chen \textit{et al.} \cite{chen2023unit3d} propose to jointly train a 3D captioner and a 3D grounder to provide diverse query descriptions for better supervising the grounding module with mutual information learning.
Based on this joint framework, Chen \textit{et al.} \cite{chen2022d} further incorporate a teacher-student architecture into the model design to train a robust grounder by mimicking the strong ability of the joint framework.
Liu \textit{et al.} \cite{liu2024cross} propose to utilize few-shot annotations (\textit{e.g.}, 10\% data) to train the grounding models with full supervision. Specifically, they develop a teacher-student framework with an additional captioning module to generate and correct pseudo labels for training.
Zhu \textit{et al.} \cite{zhu20233d} propose a pre-training strategy with transformers for 3D vision and text alignment, which utilizes self-attention layers for both single-modal modeling and multi-modal fusion without any sophisticated task-specific design. Specifically, it collects diverse prompts, scene graphs, 3D scans, and objects with self-supervised pre-training.
Jin \textit{et al.} \cite{jin2023context} also propose a 3D vision-language pertaining strategy with context-aware spatial-semantic alignment and mutual 3D-language masked modeling to learn more representative features. It shows great performance on the 3D grounding task.
Unal \textit{et al.} \cite{unal2023three} develop three novel stand-alone modules to improve grounding performance for challenging repetitive instances, \textit{i.e.}, instances with distractors of the same semantic class.
Two new datasets \cite{abdelreheem2024scanents3d,jia2024sceneverse} are proposed to scale the knowledge learning-ability of T-3DVG. 

\begin{figure}[t!]
\centering
\includegraphics[width=1.0\columnwidth]{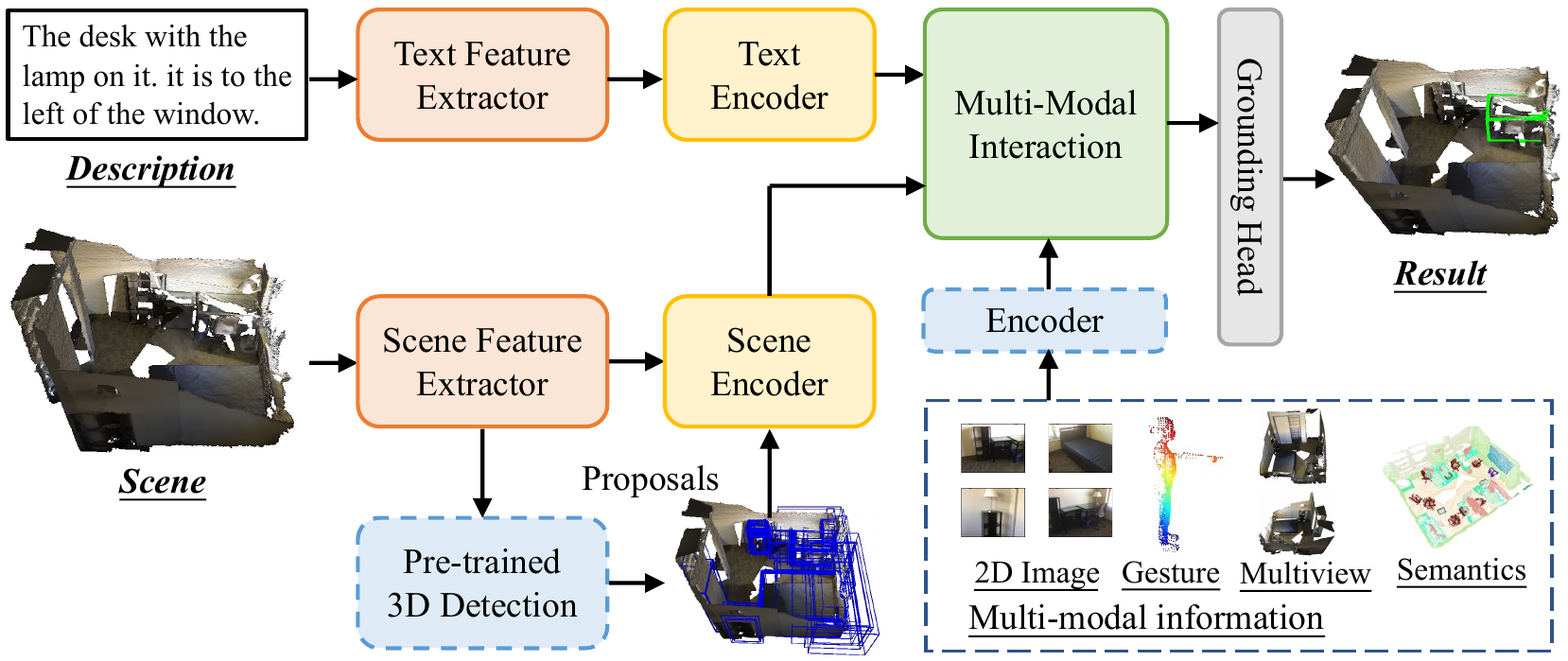}
\caption{A general pipeline for multi-modal enhanced proposal-then-selection methods.}
    \label{fig:method3}
\end{figure}

\noindent \textbf{Multi-modal enhanced proposal-then-selection methods.} 
Although the basic proposal-then-selection methods achieve significant performance, they solely exploit the contexts within the original 3D point cloud input for retrieving the specific objects according to different queries, lacking sufficient yet complementary information from other views or modalities for enriching the retrieval knowledge. To this end, there are many recent methods \cite{yang2021sat,bakr2022look,miyanishi2023cross3dvg,zhu2024unifying,xiao2024secg,lu2024scaneru,huang2022multi,guo2023viewrefer,tan2023reimagining,xu2024multi,shi2024aware} proposed to further perceive more contextual information from other perspectives (such as 2D images, multi-view contexts, \textit{etc.}) for enriching the traditional 3D grounding knowledge, as shown in Fig.~\ref{fig:method3}.

Considering the traditional point cloud data are sparse, noisy, and contain limited semantic information compared with 2D images, many works \cite{yang2021sat,bakr2022look,miyanishi2023cross3dvg,zhu2024unifying,tan2023reimagining,zhang2024towards,xu2024multi} propose to utilize additional 2D image input to assist 3D visual grounding. These methods generally first detect potential object proposals through 3D object detector, then encode each object's 2D image into feature vectors to combine with 3D object features for aligning with the semantics of sentence description. Since more information is utilized, these works achieve better performance than basic proposal-then-selection frameworks.
In addition to the point cloud input, Xiao \textit{et al.} \cite{xiao2024secg} propose to build a new representation for 3D points in the semantic point cloud defined on high-level semantics. This new representation guides the model to understand more relationships by simplified representation of objects, rather than limited to traditional appearance learning.
Both semantic and appearance object features are concatenated to feed into the latter cross-modal reasoning module.
To make T-3DVG more practical in real-world applications, Lu \textit{et al.} \cite{lu2024scaneru} introduce a human-robot interaction fashion that takes human guidance as a cue to facilitate the development of 3D visual grounding.
A point cloud of a human agent is specially introduced as additional input along with a designed gesture extraction module. Therefore, the object-level scene representations are reasoned with both human guidance and textual semantics to determine the location of target object.
Besides, multi-view learning designs \cite{huang2022multi,guo2023viewrefer,shi2024aware} is also a promising multi-modal enhancement strategy. They take multi-view point clouds of the same 3D scene as inputs, and mine their correspondences between the same object for grasping and comprehending the view knowledge. 
Overall, these multi-modal enhanced two-stage methods achieve better performance as more contextual information is embedded into the representation learning.

\subsubsection{One-Stage Methods}
Without relying on the quality of pre-trained object generators (\textit{i.e.}, 3D detector or segmentor), some recent T-3DVG methods follow the one-stage framework that trains the 3D grounding models end-to-end from the feature extraction to the final cross-modal grounding. 
Among them, inspired by the already successful two-stage approach, some works \cite{huang2023dense,yin2024lamm,li2023uni3dl,feng2024pointcloud,islam2023patron,zhang2024vision,yuan2023visual,yuan2022toward,wu20243d} propose a unified detection-interaction method that jointly trains both object detection and object reasoning with mutual learning.
To reduce the training parameters and resources, recent works \cite{luo20223d,wu2023eda,wang20233drp,lin2023unified,jain2022bottom,wang2024g} directly encode the point-wise features of the whole point cloud and globally match all points with the query to regress the optimal locations of object boundary box locations.

\begin{figure}[t!]
\centering
\includegraphics[width=1.0\columnwidth]{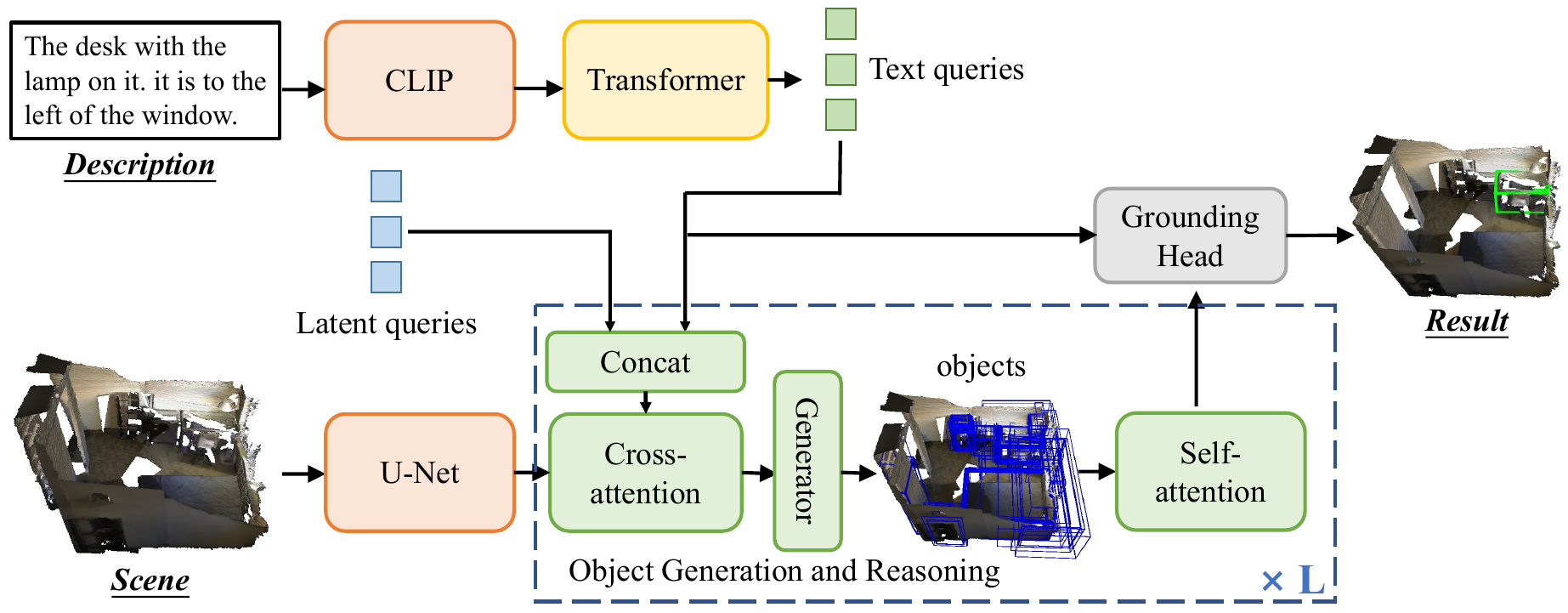}
\caption{Uni3DL architecture, reproduced from Li \textit{et al.} \cite{li2023uni3dl}.}
    \label{fig:uni3dl}
\end{figure}

\noindent \textbf{Unified detection-interaction methods.} 
To build an end-to-end yet unified detection-interaction framework, 
Li \textit{et al.} \cite{li2023uni3dl} introduce a query transformer to learn task-agnostic semantics via U-Net \cite{choy20194d} and bbox/mask outputs by attending to 3D point-level visual features, as illustrated in Fig.~\ref{fig:uni3dl}. In particular, its proposed Uni3DL learns cross- and self-attended contexts from latent and text queries, and also aggregates additional information from different object-level datasets and multi-task learning to improve the grounding performance.
Yuan \textit{et al.} \cite{yuan2022toward} also employ transformer-architecture to achieve unified detection-interaction 3D grounding.

Based on Uni3DL, RoMa \cite{feng2024pointcloud} and 3DVLP \cite{zhang2024vision} further utilize a contrastive learning strategy in the vision-language pertaining process to distinguish the text-matched and text-mismatched object pairs. Their frameworks are able to learn more representative object proposals from the transformer-based encoder. Islam \textit{et al.} \cite{islam2023patron} leverages the user's behavior as additional multimodal cues to assist the object grounding.
Yuan \textit{et al.} \cite{yuan2023visual} deploy the one-stage grounding framework in a more challenging zero-shot setting. Unlike the other methods, it only has scene data as input. To construct text descriptions for supervision, this paper utilizes language models with human interaction to generate potential sentences corresponding to each object. Then, it develops an attention-based model to jointly detect and reason text-related objects for grounding.
Wu \textit{et al.} \cite{wu20243d} propose an end-to-end superpoint-text catching mechanism, which directly correlates linguistic indications with their respective superpoints, clusters of semantically related points. It can efficiently harness cross-modal semantic relationships, primarily leveraging densely annotated superpoint-text pairs, elevating both the localization and segmentation capacities.
Although the above unified detection-interaction methods have achieved great performance, their models are heavy and require large parameters and resource costs.

\begin{figure}[t!]
\centering
\includegraphics[width=1.0\columnwidth]{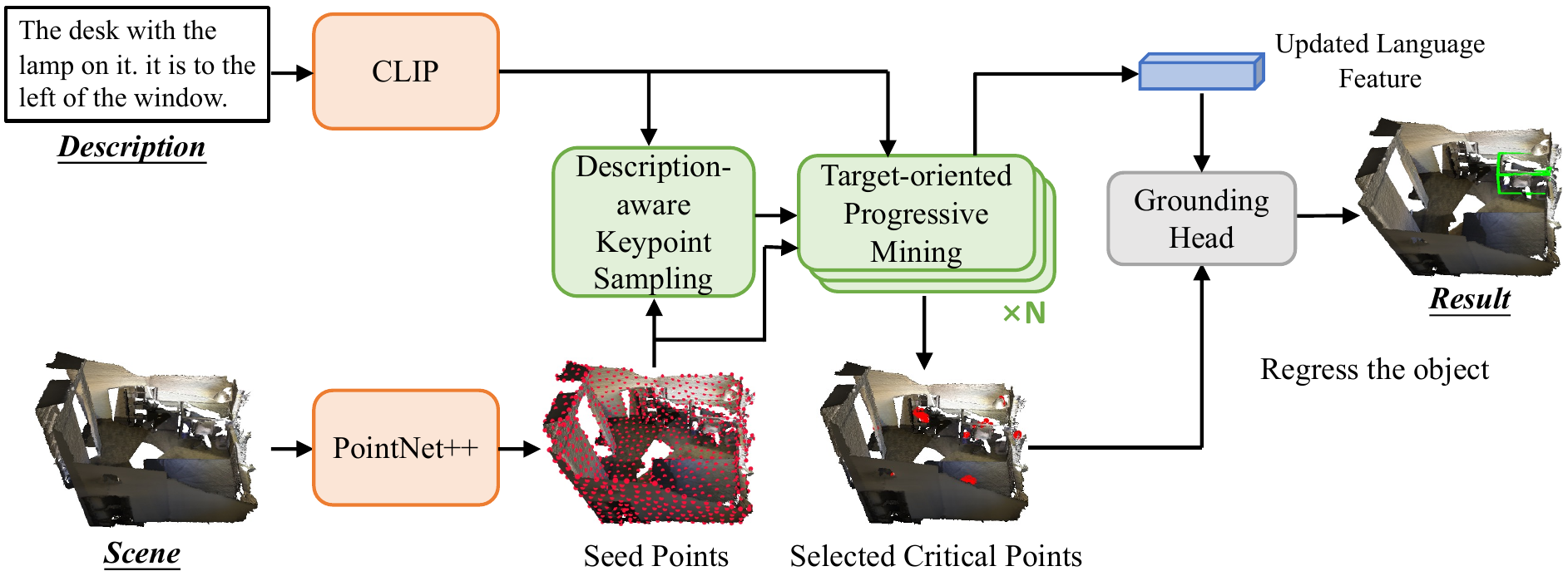}
\caption{3D-SPS architecture, reproduced from Luo \textit{et al.} \cite{luo20223d}.}
    \label{fig:sps}
\end{figure}

\noindent \textbf{Efficient regression-based methods.}
Instead of relying on any bbox proposal generation strategy, Luo \textit{et al.} propose a pioneering grounding model named 3D-SPS \cite{luo20223d} to directly regress the bbox of target object as output, as shown in Fig.~\ref{fig:sps}. By encoding the coarse-grained point-level features from pre-trained PointNet++ \cite{qi2017pointnet++}, they first coarsely interact textual semantics with each point to filter out text-irrelevant points. Then, a more fine-grained cross-modal reasoning module is utilized to find the critical points that assist in the final grounding. At last, regression heads are exploited to generate high-quality bbox results on representations of these points. Compared to previous detection-based frameworks, this model is more efficient as it does not rely on the complicated reasoning process among multiple object proposals.
Following 3D-SPS, the EDA method \cite{wu2023eda} is further proposed. It claims that previous works simply extract the sentence-level features for cross-modal alignment, losing more contextual word-level information and neglecting other attributes. Therefore, to make a more fine-grained word-level cross-modal matching, EDA proposes a text decoupling module to produce textual features for every semantic component. A position-aware and semantic-aware word-level context loss is further developed to supervise the object regression. Since EDA aggregates more granularities' context, it significantly improves the grounding performance, even much higher than most existing two-stage frameworks. This method is also flexible and can utilize additional bbox location knowledge for training.

Based on the powerful grounding ability of 3D-SPS and EDA, the latter works \cite{wang20233drp,lin2023unified,qian2024multi} focus more on the detailed network designs.
Wang \textit{et al.} and Jain \textit{et al.} \cite{wang20233drp,jain2022bottom} emphasize the importance of relative spatial relationships among objects described by the sentence description, and propose a 3D relative position module to learn and reason the spatial relations between the key point features before regressing their corresponding bbox outputs.
Huang \textit{et al.} \cite{huang2023dense} develop a contextual global transformer with learnable queries to iteratively generate text-described objects in the 3D scene. Then, it develops a fine-grained local transformer to associate the object features for scoring and selection.
This paper designs more efficient regression module within the grounding pipeline, achieving end-to-end training and reducing the resource costs. Moreover, unlike previous pre-set objects containing lots of background objects, the generated objects of this paper are text-related, significantly improving the bbox quality.
Lin \textit{et al.} \cite{lin2023unified} incorporate the superpoint mask segmentation into the object grounding task with a mutual learning framework for joint training, bringing additional mask-level knowledge for improving the grounding performance. 
Wang \textit{et al.} \cite{wang2024g} design contextual geometric feature modeling and complex utterance understanding with further semantic-geometric consistency learning, providing sufficient knowledge for regressing the object bbox.
Compared to unified detection-interaction methods, these regression-based methods are more efficient.

\begin{figure}[t!]
\centering
\includegraphics[width=1.0\columnwidth]{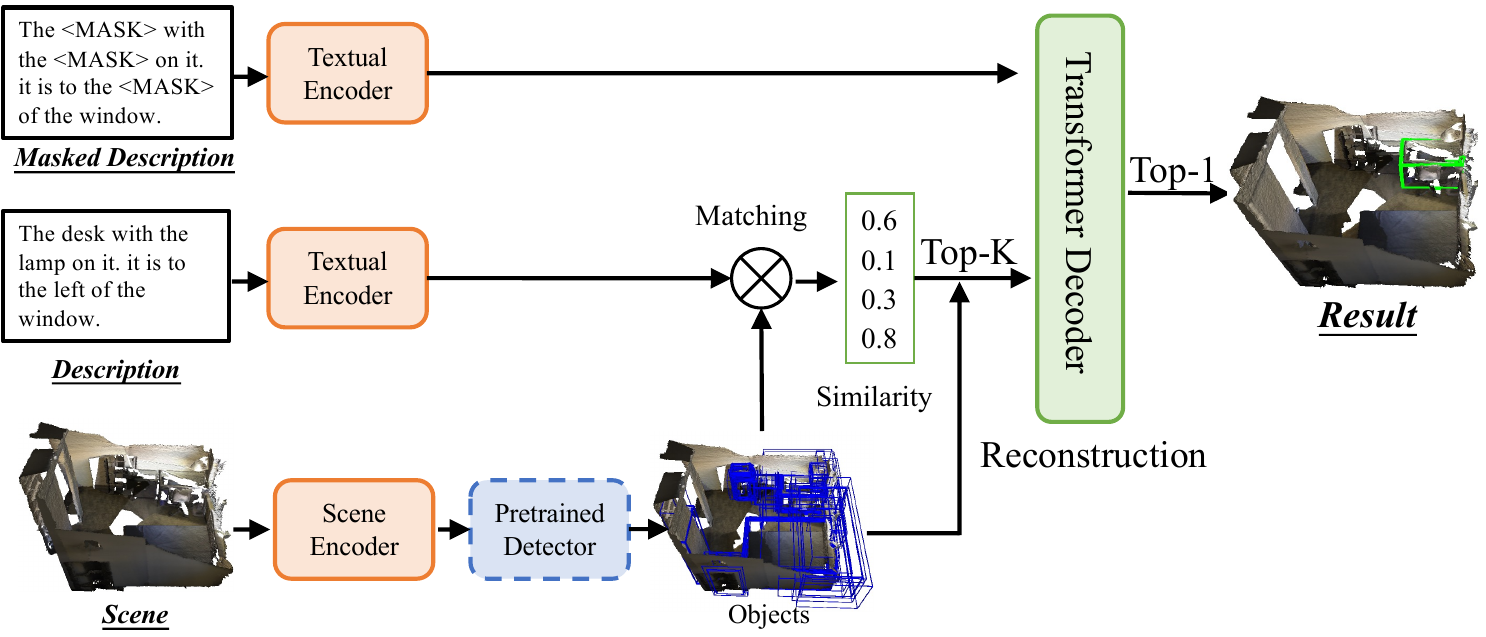}
\caption{WS-3DVG architecture, reproduced from Wang \textit{et al}. \cite{wang2023distilling}.}
    \label{fig:ws}
\end{figure}

\subsection{Weakly-Supervised Methods}
Considering the above fully-supervised methods severely rely on a large amount of training annotations, a few recent works \cite{wang2023distilling,xu2023weakly} try to explore the T-3DVG methods in a weakly supervised setting.
Instead of using complicated query-object-scene annotations for supervision, these methods solely have the knowledge of the relationship between scenes and queries without object-level bounding boxes.

Wang \textit{et al.} \cite{wang2023distilling} propose the first weakly-supervised framework for T-3DVG task, as shown in Fig.~\ref{fig:ws}.
To fill the information gap between 3D visual features and textual features, it first extracts all possible objects from the whole 3D scene with a pre-trained 3D detector. Then, it coarsely selects the top-$K$ object proposals based on their semantic similarities with the sentence description. 
To further mine their potential relationships, it borrows the text-based pre-training strategies from transformer \cite{vaswani2017attention}, which masks crucial words in the sentence and reconstructs the masked keywords using each selected proposal one by one. The reconstructed accuracy finely reflects the semantic similarity of each object to the query. Therefore, the best matched object is chosen as the final grounding result.
Xu \textit{et al.} \cite{xu2023weakly} also introduce a 3D-VLA network to address the weakly-supervised T-3DVG task. It leverages additional 2D semantics to assist the grounding process. Specifically, it encodes object-level features with pre-trained 3D detector and image-text features with CLIP \cite{radford2021learning}. Then, it interacts 3D-level, image-level, and text-level semantics to determine which object is matched with the textual description.
In summary, there are a few works that make attempts to investigate the challenging weakly-supervised T-3DVG problem. However, their frameworks are still limited without special designs of robust representation learning, resulting in worse performance than fully-supervised methods.

\subsection{Methods with Other-Modality Input} 
Instead of using a point cloud as the 3D grounding input, there are also a few works \cite{liu2021refer,wang2023embodiedscan,zhan2024mono3dvg} proposed to address the task from a new perspective with other types of inputs, such as RGB-D images and human-interaction vectors.
Liu \textit{et al.} \cite{liu2021refer} propose to ground objects in single-view RGB-D images. Given an RGB-D image and a sentence query input, they first fuse the multi-modal features via attention mechanisms and generate a visual heatmap to coarsely localize the relevant regions in the RGB-D image. Then, an adaptive feature learning based on the heatmap is introduced to perform the object-level matching to finely locate the referred object. Different from previous point-cloud-based grounding, this RGBD-based grounding requires fewer resources to collect and annotate data. Another RGBD-based method \cite{wang2023embodiedscan} is also proposed to handle the holistic multi-modal 3D perception suite. In addition to the RGB-D input, this method also feeds corresponding RGB images into the model. The whole network encodes both 2D and 3D visual features to densely interact with sentence queries, leading to more fine-grained object grounding.
Instead of relying on RGB-D data input, Zhan \textit{et al.} \cite{zhan2024mono3dvg} propose to ground target objects in monocular images. As we know, the task of monocular object detection \cite{chen2016monocular,simonelli2019disentangling} task is well developed and has a large amount of collected data, providing more robust training.
Besides, PATRON \cite{islam2023patron} takes multiple views of non-verbal interactions to locate the target objects in more practical robotics scenarios.
Lin \textit{et al.} \cite{lin2023wildrefer} also introduce a wild case of automatic driving data and corresponding 2D images for grounding.
Overall, the above methods provide a new perspective for T-3DVG instead of understanding and reasoning point clouds.

\begin{figure}[t!]
\centering
\includegraphics[width=1.0\columnwidth]{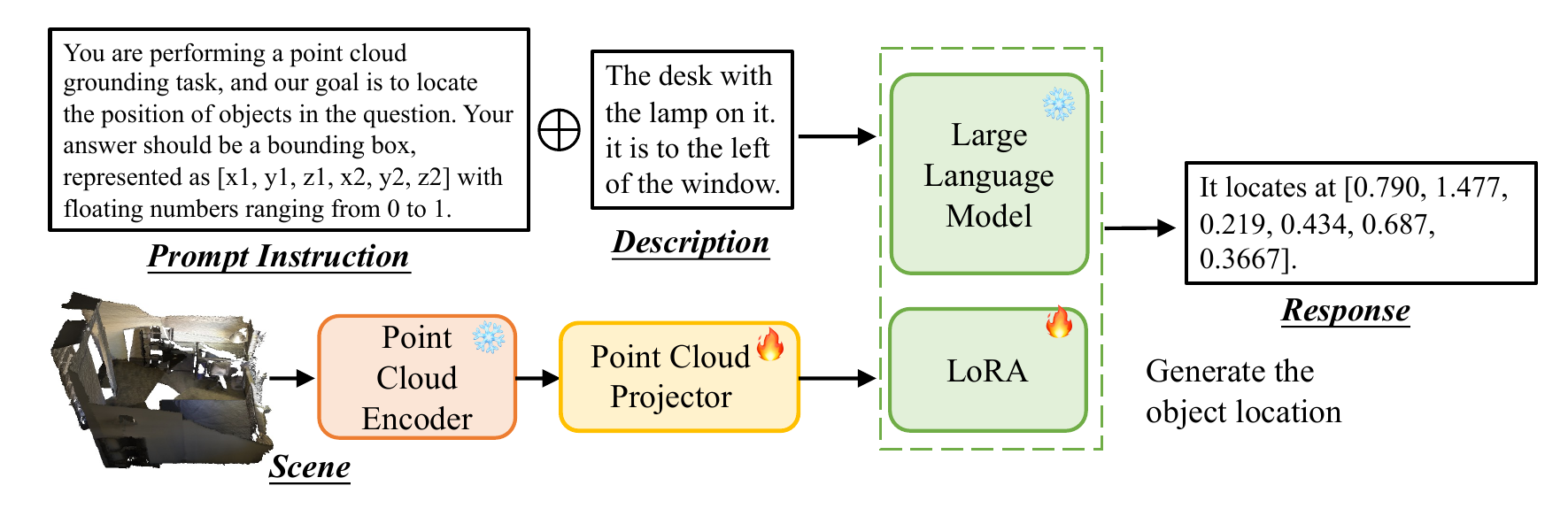}
\caption{LAMM architecture, reproduced from Yin \textit{et al}. \cite{yin2024lamm}.}
    \label{fig:lamm}
\end{figure}

\subsection{Methods using Large Models}
By benefiting from the strong comprehension of large language models (LLMs), there are many recent T-3DVG methods \cite{yang2023llm,zhan2024mono3dvg,guo2023viewrefer,yuan2023visual,fang2023transcribe3d,bakr2023cot3dref,wu2024dora,yin2024lamm,li20243dmit,jia2024sceneverse,feng2024naturally,yang20243dgrand,zhu2024empowering} that adopt large models to construct their grounding framework.
Considering the description annotations are complicated but with limited information,
most works \cite{yang2023llm,zhan2024mono3dvg,guo2023viewrefer,yuan2023visual} utilize LLM tools \cite{yang2023auto,schick2024toolformer,wu2023brief,radford2019language,feng2024naturally} to decompose complex natural language queries into semantic constituents for adaptive grounding, and enrich the contexts of original description with designed prompts for providing more detailed commonsense knowledge, or directly generate a new description from another perspective via captioners.
Yang \textit{et al.} \cite{yang2023llm} further utilize LLMs to evaluate the spatial relations among the proposed objects to make a final grounding decision.
These methods incorporated more diverse textual semantics into the text-object reasoning process, leading to more precise grounding results.
Besides, some works \cite{fang2023transcribe3d,bakr2023cot3dref,wu2024dora} attempt to leverage the commonsense reasoning capability of LLMs to make inferences about the objects in the scene and their interactions. Specifically, Bakr \textit{et al.} \cite{bakr2023cot3dref} and Wu \textit{et al.} \cite{wu2024dora} propose to exploit Chain-of-Thoughts \cite{kojima2022large,wei2022chain} to decompose the referring task into multiple interpretable steps, whereas to reach the final target the model must first predict the anchors one by one, in a logical order.
However, the above methods simply take the LLMs as a module tool to assist the traditional grounding model, failing to exploit the strong power of the LLM architecture designs.
To this end, recent methods \cite{yin2024lamm,li20243dmit,jia2024sceneverse,zhu2024empowering} extend the reasoning ability of multi-modal LLMs \cite{liu2024visual,zhu2023minigpt,ye2023mplug} into the 3D scenarios, as shown in Fig.~\ref{fig:lamm}. Specifically, they design object or scene-level 3D encoders to embed the visual input into the same latent vectors as the language dimensions in LLMs. Then, both textual and visual embeddings are jointly fed into the LLMs for reasoning. The grounding results can be achieved by inputting a specific prompt into the LLMs.

\subsection{Discussion}
In summary, existing T-3DVG methods can be generally categorized into two types: fully-supervised approaches and weakly-supervised approaches.
As for fully-supervised approaches, most works follow a basic proposal-then-selection framework, which first utilizes pre-trained models to extract all possible object proposals in each scene and then match them with textual sentences for selection.
To enhance the prior knowledge for assisting grounding, some proposal-then-selection methods also utilize multi-modal data as additional input to provide sufficient grounding information.
Although this two-stage framework achieves significant performance, it severely relies on the quality of the pre-trained model and costs large resources to compute background objects.
Therefore, one-stage methods are proposed to directly regress the target objects. Some of them follow the two-stage methods to design a unified detection-interaction frameworks, which replace the pre-trained object model with an efficient learnable one. The others directly regress the target object with point-level feature reasoning.
As for weakly-supervised approaches, they solely exploit the knowledge of the relationship annotations between scenes and queries without object-level bounding boxes. Existing weakly-supervised works generally follow a proposal-then-matching framework that scores potential proposals with self-learning strategies. 
There are also some works that utilize other 3D-modal input instead of point clouds, or combine LLMs with traditional grounding models, or propose to address outdoor scenes \cite{liu2024talk,guan2024talk2radar}.
Overall, there is still plenty of room for exploration in developing methods of weakly-supervised approaches and methods of using large models.

\section{Datasets and Measures}
\subsection{Benchmark Datasets}

\begin{figure*}[!tbp]
\centering
\includegraphics[width=1.85\columnwidth]{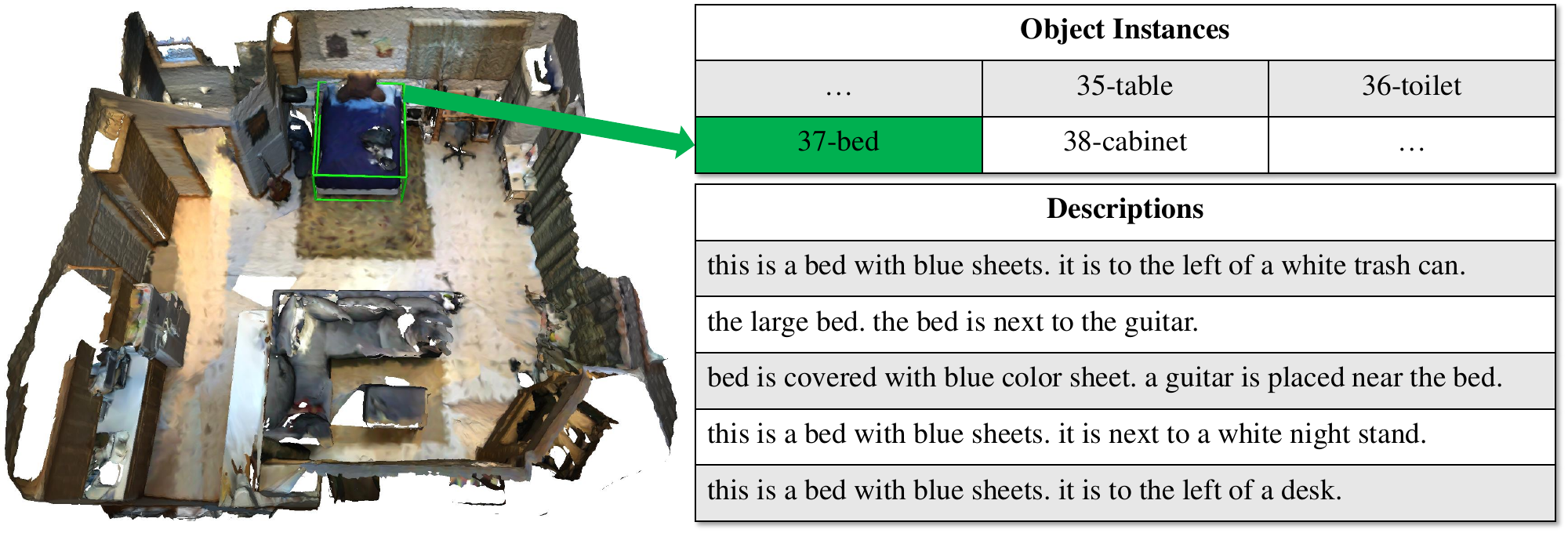}
\caption{Depiction of a typical instance from scene\_id 0000\_00 within the ScanRefer Dataset \cite{chen2020scanrefer}. This dataset is recognized for its thorough object labeling methodology, where almost every object in the scene is annotated and provided with five associated descriptions.}
\label{fig:figScanrefer}
\end{figure*}

Achieving optimal performance in the T-3DVG task involving 3D vision and language heavily relies on the availability of extensive datasets. However, the process of acquiring and annotating volumetric data presents formidable hurdles, resulting in smaller collections of RGB-D datasets \cite{shotton2013scene,xiao2013sun3d,song2016deep,hua2016scenenn} for 3D scenarios compared to their 2D counterparts. Consequently, researchers often resort to employing synthetic data to supplement the dearth of authentic data from the real world \cite{wu20153d,chang2015shapenet}. To tackle this challenge, Dai \textit{et al.} pioneered the creation of the ScanNet dataset \cite{dai2017scannet}, an extensively annotated compilation of 3D indoor scans derived from real-world settings, captured through a proprietary RGB-D capture mechanism.

Comprising 2.5 million perspectives derived from 1513 scans, drawn from 707 unique environments, the ScanNet dataset surpasses other well-known datasets \cite{silberman2012indoor,sturm2012benchmark,xiao2013sun3d,savva2016pigraphs,hua2016scenenn} like NYUv2 \cite{silberman2012indoor}, SUN3D \cite{xiao2013sun3d}, and SceneNN \cite{hua2016scenenn} in terms of scale. Nevertheless, it is essential to recognize that ScanNet primarily caters to tasks such as categorizing 3D objects, labeling semantic voxels, and retrieving 3D objects with annotations at the instance level within a 3D point cloud. Consequently, it may not be directly applicable to tasks like 3D visual grounding and 3D dense captioning. To address this constraint, novel datasets such as ScanRefer \cite{chen2020scanrefer}, Nr3D \cite{achlioptas2020referit3d}, and Sr3D \cite{achlioptas2020referit3d} have emerged, specifically tailored for 3D visual grounding tasks, building upon the foundation laid by ScanNet. Both ScanRefer and Nr3D/Sr3D curate subsets of scenes from ScanNet, accompanied by additional manually crafted descriptions for the objects within each scene. Moreover, they provide accessible online platforms for data visualization, rendering them invaluable assets for the development and assessment of models in the realm of 3D visual grounding tasks.

\begin{table}[!t]
\centering
\caption{The statistical attributes of the ScanRefer, Nr3D and Sr3D datasets. These benchmarks predominantly feature 3D scenes, object categories, and meticulously crafted descriptive annotations. However, discernible disparities emerge in their annotation methodologies. ScanRefer opts for comprehensive labeling across all objects within a scene, whereas Nr3D and Sr3D prioritize the annotation of selected categories that exhibit elevated occurrence frequencies.}
\vspace{-6pt}
\begin{tabular}{c|c|c|c}
\hline
Number \#                      & ScanRefer & Nr3D   & Sr3D   \\ \hline
Descriptions                   & 51,583    & 41,503 & 83,572 \\ \hline
Scenes                         & 800       & 707    & 1273 \\ \hline
Objects / Contexts             & 11,046    & 5,878  & 11,375 \\ \hline
Object classes                 & 265       & 76     & 100 \\ \hline
Objects / Contexts per scene   & 13.81     & 8.31   & 8.94 \\ \hline
Descriptions per object        & 4.67      & 7      & 9.43 \\ \hline
Average length of descriptions & 20.27     & 11.4   & 9.68 \\ \hline
\end{tabular}
\label{tab:datasetStats}
\end{table}

\begin{figure}[!tbp]
\centering
\includegraphics[width=0.99\columnwidth]{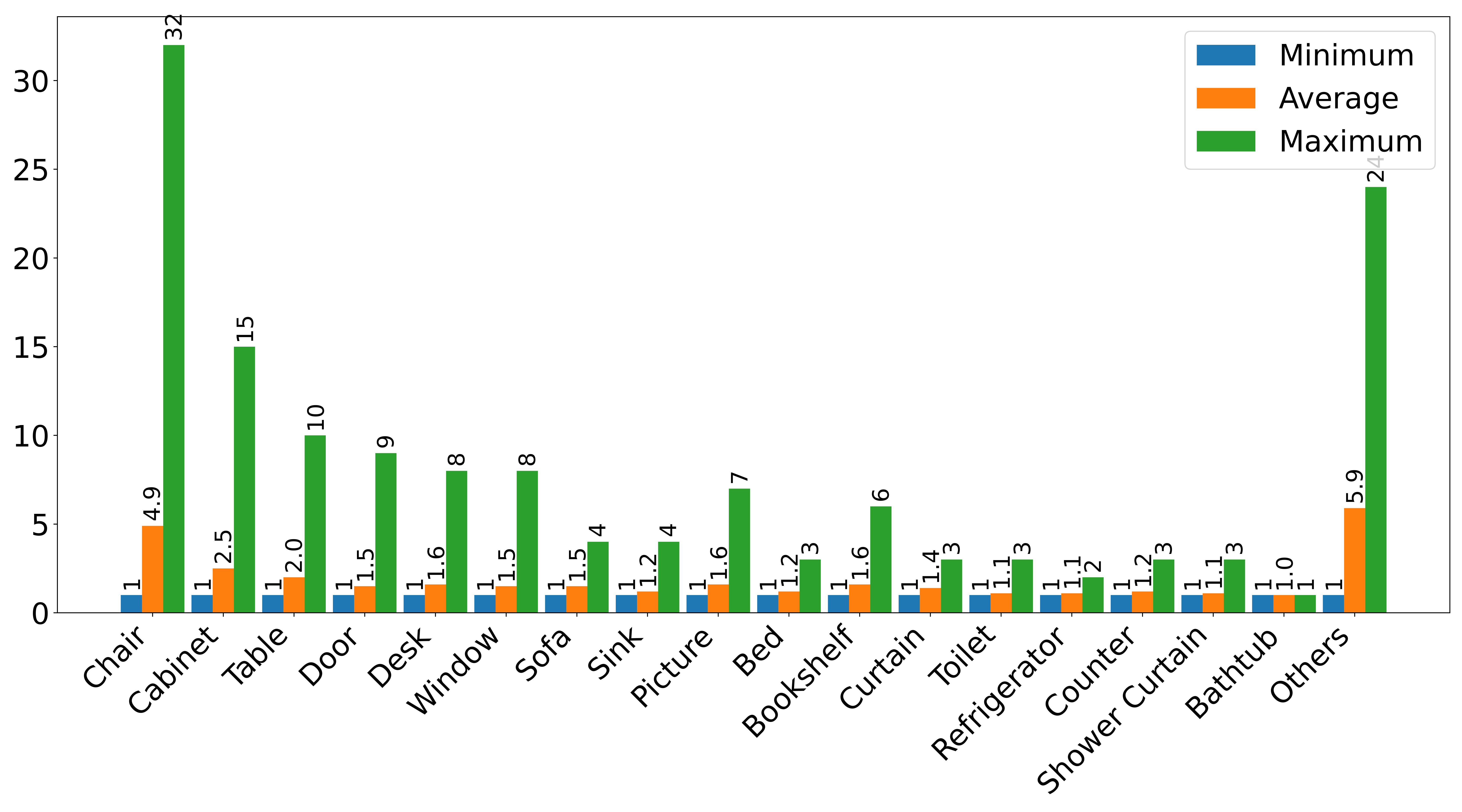}
\vspace{-16pt}
\caption{Demonstration of the statistics of 18 general categories, consisting of 256 sub-categories within a 3D indoor environment, predominantly featuring furniture items like chairs, tables, and cabinets. The terms "Maximum," "Average," and "Minimum" signify the highest, mean, and lowest counts, respectively, for a specific type of object present in a scene. For instance, a scene may contain as many as 32 chairs.}
\label{fig:scanreferStat}
\end{figure}

\noindent \textbf{ScanRefer.} ScanRefer stands out as a prominent dataset tailored for tasks in 3D visual grounding, furnishing a significant array of natural language descriptions for objects found within the scans sourced from the ScanNet dataset \cite{dai2017scannet}. In terms of statistics, this dataset encompasses 51,583 elaborate and varied descriptions corresponding to 11,046 objects across 800 scans from ScanNet. These descriptions span more than 250 categories of common indoor objects, detailing attributes such as hue, dimensions, morphology, and spatial arrangements. Each object within a scene receives roughly five manual annotations, contributing to the dataset's richness and comprehensiveness. The comprehensive statistical breakdown of ScanRefer can be found in the secondary column of Table \ref{tab:datasetStats}. An exemplary depiction of a typical scene from ScanRefer, specifically scene 0000\_00, is depicted in Fig. \ref{fig:figScanrefer}. Following ScanNet's official split \cite{dai2017scannet}, the dataset is divided quantitatively into training, validation, and test subsets, comprising 36,665, 9,508, and 5,410 instances, respectively. Given the absence of an officially disclosed unseen test split, most analyses are conducted on the validation subset \cite{chen2020scanrefer}. Table \ref{tab:ScanReferStats} offers insights into the distribution statistics of the ScanRefer dataset. Moreover, Fig. \ref{fig:scanreferStat} visually elucidates the frequency distribution of several prominent object categories within ScanRefer.

\begin{table}[!t]
\centering
\caption{The standard split's statistics of the ScanRefer dataset.}
\vspace{-6pt}
\begin{tabular}{c|c|c|c|c}
\hline
Number \#                      & Train & Val   & Test & Total   \\ \hline
Descriptions                   & 36,665 & 9,508 &5,410 &51,583 \\ \hline
Scenes                         & 562 &141 &97 &800 \\ \hline
Objects             & 7,875 &2,068 &1,103 &11,046 \\ \hline
Objects per scene   & 14.01 &14.67 &11.37 &14.14 \\ \hline
Descriptions per scene        & 65.24 &67.43 &55.77 &65.68 \\ \hline
Descriptions per object & 4.66 &4.60& 4.90& 4.64 \\ \hline
\end{tabular}
\label{tab:ScanReferStats}
\end{table}

\begin{figure*}[!tbp]
\centering
\includegraphics[width=1.85\columnwidth]{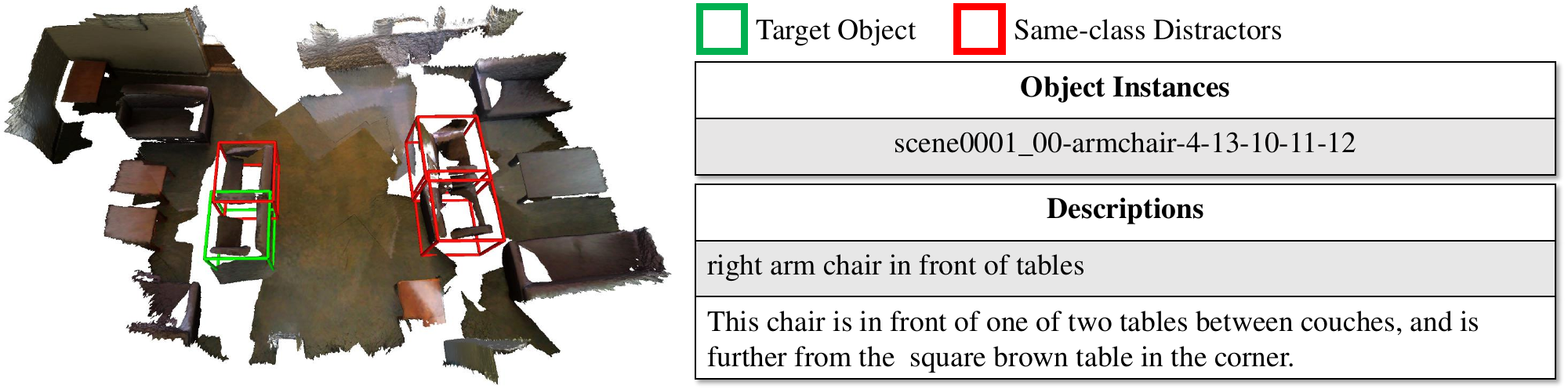}
\caption{Depiction of a standard example from scene 0001\_00 in the Nr3D dataset \cite{achlioptas2020referit3d}. The Nr3D dataset focuses on selective annotation of frequently occurring object categories, primarily involving multiple objects, with each one linked to several distinct descriptions.}
\label{fig:figNr3d}
\end{figure*}

\noindent \textbf{Nr3D.} The 3D Natural Reference (Nr3D) dataset constitutes a pivotal component of the ReferIt3D collection \cite{achlioptas2020referit3d}, alongside the synthetic language descriptions dataset Sr3D \cite{achlioptas2020referit3d}. Illustrated in Fig. \ref{fig:figNr3d} is a typical portrayal of scene 0001\_00 sourced from Nr3D. Nr3D is specifically tailored to address fine-grained objects within three-dimensional space, thereby presenting heightened complexity compared to ScanRefer, where the reference object shares similar characteristics \cite{chen2022language}. Comprising 41,503 spontaneous, unrestricted verbal expressions delineating objects across 76 meticulously defined object categories within 5,878 communicative contexts, Nr3D encapsulates a nuanced understanding of object semantics. These communicative contexts are articulated as distinct sets denoted by {S,C}, wherein S denotes one of the 707 unique ScanNet scenes, and C signifies the fine-grained classification of S, thus encapsulating instances of the same object type within a scene. Analogous to ScanRefer, Nr3D adheres to the established ScanNet partitions for experimentation purposes. A comprehensive breakdown of Nr3D's statistics is provided in the adjacent column of Table \ref{tab:datasetStats}.

\noindent \textbf{Sr3D.} The dataset known as Sr3D (Spatial Reference in 3D) is a component of the ReferIt3D collection \cite{achlioptas2020referit3d}, encompassing a total of 83,572 expressions. Each expression serves the purpose of uniquely identifying a particular object within a ScanNet 3D scene by establishing a connection between the target and a neighboring object (referred to as an anchor). Anchors represent instances of objects drawn from a pool of 100 distinct classes within ScanNet. However, it's crucial to note that an anchor never shares the same class as the target or its distractors. To illustrate, consider a hypothetical 3D scene containing a target object (such as a desk) which can be definitively distinguished from other objects through a spatial relationship (e.g., closest) to an anchor object (e.g., door). Discriminative expressions in Sr3D are constructed using a compositional template comprised of three placeholders: "$\langle$ target-class $\rangle$ $\langle$spatial-relation$\rangle$ $\langle$anchor-class(es)$\rangle$". For instance, "the desk that is closest to the door". Five distinct types of spatial relationships between objects are delineated: "horizontal proximity", "vertical proximity", "between", "allocentric", and "support". A detailed breakdown of Sr3D's characteristics is presented in the final column of Table \ref{tab:datasetStats}.

\begin{figure}[!tbp]
\centering
\includegraphics[width=0.99\columnwidth]{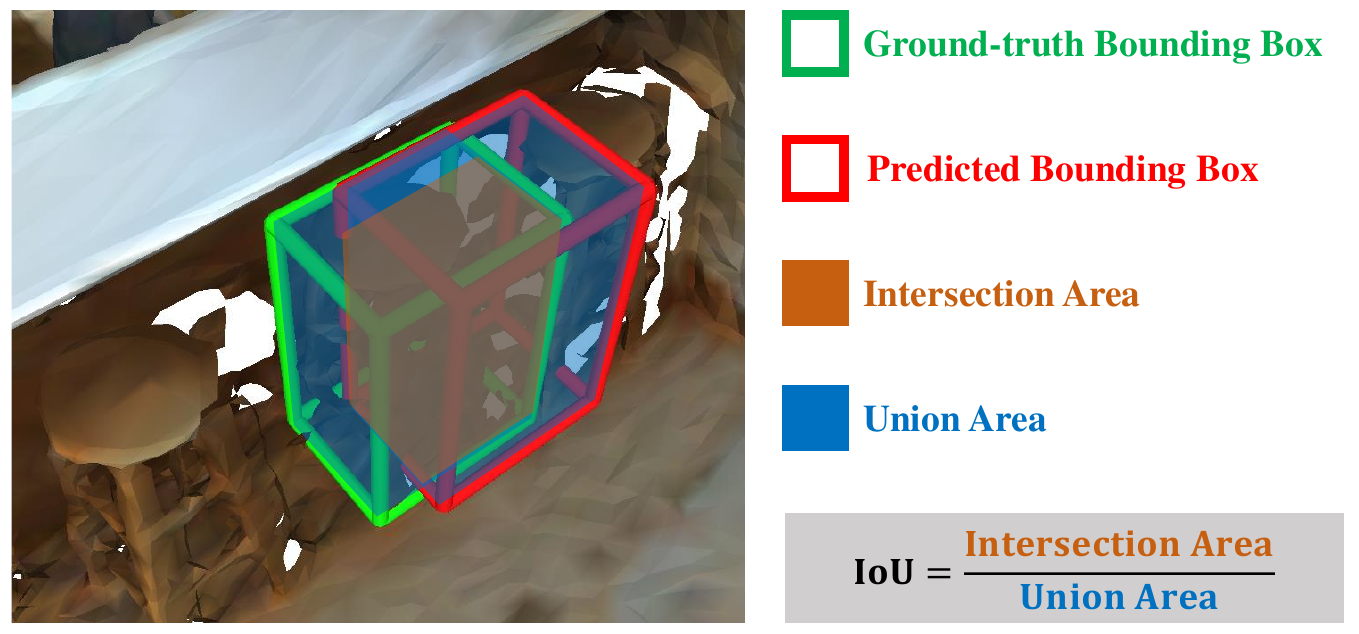}
\caption{An illustration of the 3D IoU metric.}
\label{fig:figIou}
\end{figure}

\begin{table*}[t!]
\centering
\caption{Performance Comparison on ScanRefer dataset. '*' denotes training with few-shot annotations (\textit{i.e.}, 10\% data).}
\vspace{-10pt}
\setlength{\tabcolsep}{1.0mm}{
\begin{tabular}{ccccc|cccccc}
\hline
\multirow{2}{*}{Method}                                        & \multirow{2}{*}{Venue}     & \multirow{2}{*}{Supervise} & \multirow{2}{*}{Category}  & \multirow{2}{*}{Modality} & \multicolumn{2}{c}{Unique} & \multicolumn{2}{c}{Multiple}     & \multicolumn{2}{c}{Overall}     \\
                                                               &                            &                            &                            &                           & Acc@0.25       & Acc@0.5      & Acc@0.25 & Acc@0.5& Acc@0.25 & Acc@0.5 \\ \hline \hline
ScanRefer \cite{chen2020scanrefer}                                                      & ECCV 2020                  & fully                       & two-stage                  & 3D                       & 67.64&46.19&32.06&21.26&38.97          & 26.10        \\
Referit3D \cite{achlioptas2020referit3d}                                                     & ECCV 2020                  & fully                       & two-stage                  & 3D                        &53.75&37.47&21.03&12.83& 26.44          & 16.90             \\
TGNN   \cite{huang2021text}                                                        & AAAI 2021                  & fully                       & two-stage                  & 3D                        &68.61&56.80&29.84&23.18& 37.37          & 29.70        \\
3DVG-Transformer \cite{zhao20213dvg}                                                & ICCV 2021                    & fully                       & two-stage                  & 3D                        &77.16&58.47&38.38&28.70& 45.90          & 34.47        \\
FFL-3DOG   \cite{feng2021free}                                                    & ICCV 2021                  & fully                       & two-stage                  & 3D                        &78.80&67.94&35.19&25.70& 41.33          & 34.01        \\
InstanceRefer       \cite{yuan2021instancerefer}                                           & ICCV 2021                  & fully                       & two-stage                  & 3D                        &77.45&66.83&31.27&24.77& 40.23          & 32.93        \\
3D-SPS \cite{luo20223d}                                            & CVPR 2022                & fully                       & one-stage                  & 3D                        &81.63&64.77&39.48&29.61& 47.65          & 36.43        \\
3DJCG \cite{cai20223djcg}                                           & CVPR 2022                & fully                       & two-stage                  & 3D                        &78.75&61.30&40.13&30.08& 47.62          & 36.14         \\
MVT  \cite{huang2022multi}                                                          & CVPR 2022                  & fully                       & two-stage                  & 3D                        &77.67&66.45&31.92&25.26& 40.80          & 33.26         \\
BUTD-DETR   \cite{jain2022bottom}                                                   & ECCV 2022                  & fully                       & one-stage                  & 3D                        &81.47&61.24&44.20&32.81& 49.76          & 37.05        \\

ViL3DRel     \cite{chen2022language}                                                  & NeurIPS 2022               & fully                       & two-stage                  & 3D                        &81.58&68.62&40.30&30.71& 47.94          & 37.73        \\
3D-VisTA     \cite{zhu20233d}                                                  & ICCV 2023                  & fully                       & two-stage                  & 3D                        &81.6&75.1&43.7&39.1& 50.6           & 45.8         \\

3D-VLP \cite{jin2023context}                  & CVPR 2023                  & fully                       & two-stage                 & 3D                        &79.35&62.60&42.54&32.18& 49.68          & 38.08               \\
EDA    \cite{wu2023eda}                                                  & CVPR 2023                  & fully                       & one-stage                  & 3D                       &85.76&68.57&49.13&37.64 & 54.59          & 42.26        \\

Multi3DRefer      \cite{zhang2023multi3drefer}                                                & CVPR 2023                  & fully                       & two-stage                  & 3D                       &- & 77.2&-          & 36.8        & -     & 44.7     \\
3DRP-Net    \cite{wang20233drp}                                                  & EMNLP 2023                  & fully                       & one-stage                  & 3D                        &83.13&67.74&42.14&31.95& 50.10          & 38.90        \\
CORE-3DVG \cite{yang2024exploiting}                                                   & NeurIPS 2023                  & fully                       & two-stage                  & 3D                        &84.99&67.09&51.82&39.76& 56.77          & 43.84         \\
G3-LQ \cite{wang2024g}                                                   & CVPR 2024                  & fully                       & one-stage                  & 3D                        &88.59&73.28&50.23&39.72& 55.95          & 44.72         \\
MA2TransVG \cite{xu2024multi}                                                   & CVPR 2024                  & fully                       & two-stage                  & 3D                        &86.3&74.1&53.8&41.4& 57.9          & 45.7        \\
VPP-Net \cite{shi2024aware}                                                   & CVPR 2024                  & fully                       & two-stage                  & 3D                        &86.05&67.09&50.32&39.03& 55.65          & 43.29  \\
MCLN \cite{qian2024multi} & ECCV 2024 & fully & one-stage & 3D & 84.43 & 68.36 & 49.72 & 38.41 & 54.30 & 42.64 \\
3D-CTTSF \cite{liu2024cross}* & ACMMM 2024 & fully & two-stage & 3D & 76.75 & 58.27 & 38.60 & 28.98 & 46.00 & 34.67
\\ \hline
ScanRefer \cite{chen2020scanrefer}                                                      & ECCV 2020                  & fully                       & two-stage                  & 3D+2D                       &76.33&53.51&32.73&21.11& 41.19          & 27.40        \\
3DVG-Transformer \cite{zhao20213dvg}                                                & ICCV 2021                    & fully                       & two-stage                  & 3D+2D                     &81.93&60.64&39.30&28.42& 47.57          & 34.67        \\
SAT       \cite{yang2021sat}                                                     & ICCV 2021                  & fully                       & two-stage                  & 3D+2D                     &73.21&50.83&37.64&25.16& 44.54          & 30.14       \\
3D-SPS \cite{luo20223d}                                            & CVPR 2022                & fully                       & one-stage                   & 3D+2D                     &84.12&66.72&40.32&29.82& 48.82          & 36.98       \\
3DJCG \cite{cai20223djcg}                                           & CVPR 2022                & fully                       & two-stage                  & 3D+2D                     &83.47&64.34&41.39&30.82& 49.56          & 37.33       \\
D3Net      \cite{chen2022d}                                                    & ECCV 2022                  & fully                       & two-stage                  & 3D+2D                     & -&72.04              &-& 30.05        & -        & 37.87        \\
WS-3DVG      \cite{wang2023distilling}                                                  & ICCV 2023                  & weakly                       & two-stage                  & 3D+2D                     &-&-&-&-& 27.37          & 21.96        \\

3D-VLP \cite{jin2023context}                  & CVPR 2023                  & fully                       & two-stage                 & 3D+2D                     &84.23&64.61&43.51&33.41& 51.41          & 39.46               \\ 
3DVLP \cite{zhang2024vision}                  & AAAI 2024                  & fully                       & one-stage                  & 3D+2D                     &85.18&70.04&43.65&33.40& 51.7           & 40.51              \\ 
3D-CTTSF \cite{liu2024cross}* & ACMMM 2024 & fully & two-stage & 3D+2D & 82.87 & 61.79 & 39.57 & 28.89 & 47.97 & 35.28
\\
\hline
\end{tabular}}
\vspace{-12pt}
\label{tab:perform_scanrefer}
\end{table*}

\subsection{Evaluation Metrics}
In the realm of 3D grounding assessment, there is a necessity to compare predicted locations, typically depicted as bounding boxes, against ground truth locations of objects within testing samples. Acc@$K$IoU \cite{chen2020scanrefer} stands as the predominant metric in 3D visual grounding, quantifying the proportion of positive predictions surpassing a designated intersection over union (IoU) threshold, typically denoted as $K$ and commonly set at 0.25 or 0.5. 
It is notable that certain datasets present varying evaluation configurations. For instance, in datasets like Nr3D and Sr3D, predefined object proposals are provided, with the metric focusing on the precision of selecting the target bounding box from these proposals. Additionally, various datasets assess performance across diverse scenarios. ScanRefer \cite{chen2020scanrefer} partitions datasets into unique/multiple/overall splits, while ReferIt3D \cite{achlioptas2020referit3d} categorizes them into easy/hard/view-dependent/view-independent/overall divisions. Specifically, in the ScanRefer dataset, the ``unique'' split includes instances where only a single distinct object from a specific category fits the description, while the ``multiple'' split involves cases with several objects from the same category, creating ambiguity. For example, a scene with only one refrigerator would clearly match a sentence referring to a refrigerator. On the other hand, the ReferIt3D dataset classifies ``hard'' contexts as those with more than two distractor objects, and identifies ``view-dependent'' instances where recognizing the target requires the observer to adopt a specific viewpoint within the scene. Varied methodologies are employed for assessment; some compute the average IoU \cite{hong20233d,huang2021text}, whereas others ascertain the average distance between bounding box centroids \cite{hong20233d}. 

\begin{table*}[]
\centering
\caption{Performance Comparison on Sr3D dataset. '$\dagger$' denotes evaluation with detected boxes rather than ground-truth boxes.}
\vspace{-10pt}
\begin{tabular}{ccccc|ccccc}
\hline
Method                            & Venue                      & Supervise             & Category                   & Modality  & Easy  & Hard & View-dep. & View-indep.&Overall \\ \hline \hline
Referit3D \cite{achlioptas2020referit3d}                                                     & ECCV 2020                  & fully                       & two-stage                  & 3D                        &44.7&31.5&39.2&40.8          & 40.8             \\
TGNN   \cite{huang2021text}                                                        & AAAI 2021                  & fully                       & two-stage                  & 3D                        &48.5&36.9&45.8& 45.0          & 45.0        \\
TransRefer3D   \cite{he2021transrefer3d}                                                        & MM 2021                  & fully                       & two-stage                  & 3D                        &60.5&50.2&49.9&57.7          &57.4        \\
3DVG-Transformer \cite{zhao20213dvg}                                                & ICCV 2021                    & fully                       & two-stage                  & 3D                        &54.2&44.9&44.6& 51.7          & 51.4        \\
InstanceRefer       \cite{yuan2021instancerefer}                                           & ICCV 2021                  & fully                       & two-stage                  & 3D                        &51.1&40.5&45.4& 48.1          & 48.0        \\

3DRefTransformer      \cite{abdelreheem20223dreftransformer}                                         & WACV 2022                  & fully                       & two-stage                  & 3D                        &50.7               & 38.3            & 44.3     & 47.1&47.0     \\
LanguageRefer     \cite{roh2022languagerefer}                                             & CoRL 2022                  & fully                       & two-stage                  & 3D                        & 58.9              & 49.3            & 49.2     & 56.3&56.0     \\
3D-SPS \cite{luo20223d}                                            & CVPR 2022                & fully                       & one-stage                  & 3D                        & 56.2          & 65.4        & 49.2    & 63.2&62.6     \\
MVT  \cite{huang2022multi}                                                          & CVPR 2022                  & fully                       & two-stage                  & 3D                        & 66.9          & 58.8        & 58.4     & 64.7&64.5     \\
BUTD-DETR   \cite{jain2022bottom}                                                   & ECCV 2022                  & fully                       & one-stage                  & 3D                        & -          & -        & -&-     & 65.6     \\

ViL3DRel     \cite{chen2022language}                                                  & NeurIPS 2022               & fully                       & two-stage                  & 3D           & 74.9          & 67.9        & 63.8     & 73.2&72.8                 \\
LAR   \cite{bakr2022look}                                                         & NeurIPS 2022               & fully                       & two-stage                  & 3D             & 63.0              & 51.2            & 50.0     & 59.1&59.4                \\
3D-VisTA     \cite{zhu20233d}                                                  & ICCV 2023                  & fully                       & two-stage                  & 3D                        &78.8           & 71.3         & 58.9     & 77.3&76.4     \\

EDA    \cite{wu2023eda}                                                  & CVPR 2023                  & fully                       & one-stage                  & 3D                         & -          & -        & -&-     & 68.1     \\
NS3D    \cite{hsu2023ns3d}                                                  & CVPR 2023                  & fully                       & two-stage                  & 3D                        & -          & -        & 62.0&-     & 62.7     \\
3DRP-Net    \cite{wang20233drp}                                                  & EMNLP 2023                  & fully                       & one-stage                  & 3D                        & 75.6         & 69.5        & 65.5     & 74.9&74.1     \\
CORE-3DVG \cite{yang2024exploiting}  $\dagger$                                                    & NeurIPS 2023                  & fully                       & two-stage                  & 3D & -          & -        & -&-     & 54.30                            \\
G3-LQ \cite{wang2024g}                                                      & CVPR 2024                  & fully                       & one-stage                  & 3D                         & -          & 66.3        & -&-     & 73.1    \\
MA2TransVG \cite{xu2024multi}                                                      & CVPR 2024                  & fully                       & two-stage                  & 3D                         & 76.0          & 69.3        & 64.5&73.8     & 73.9    \\
VPP-Net \cite{shi2024aware}                                                       & CVPR 2024                  & fully                       & two-stage                  & 3D                         & -          & -        & -&-     & 68.7\\
MiKASA \cite{chang2024mikasa}                                                       & CVPR 2024                  & fully                       & two-stage                  & 3D                         & 78.6          & 67.3        & 70.4&75.4     & 75.2\\ 
MCLN \cite{qian2024multi} & ECCV 2024 & fully & one-stage & 3D & - & - & - & - & 68.4 \\
GNL3D \cite{huang2024advancing} & ACMMM 2024 & fully & two-stage & 3D & 72.8 & 64.0 & 58.0 & 70.6 & 70.1 \\
\hline
SAT       \cite{yang2021sat}                                                     & ICCV 2021                  & fully                       & two-stage                  & 3D+2D                     & -          & -        & -&-     & 57.9     \\
WS-3DVG      \cite{wang2023distilling}                                                  & ICCV 2023                  & weakly                       & two-stage                  & 3D+2D                     &29.40          & 21.00        & 20.21    & 27.19&26.89    \\ \hline            
\end{tabular}
\vspace{-10pt}
\label{tab:perform_sr3d}
\end{table*}

\begin{table*}[]
\centering
\caption{Performance Comparison on Nr3D dataset. '$\dagger$' denotes evaluation with detected boxes rather than ground-truth boxes.}
\vspace{-10pt}
\begin{tabular}{ccccc|ccccc}
\hline
Method                            & Venue                      & Supervise             & Category                   & Modality  & Easy  & Hard & View-dep. & View-indep.& Overall \\ \hline \hline
Referit3D \cite{achlioptas2020referit3d}                                                     & ECCV 2020                  & fully                       & two-stage                  & 3D                        &43.6&27.9&32.5&37.1          & 35.6             \\
TGNN   \cite{huang2021text}                                                        & AAAI 2021                  & fully                       & two-stage                  & 3D                        &44.2&30.6&35.8& 38.0          & 37.3        \\
TransRefer3D   \cite{he2021transrefer3d}                                                        & MM 2021                  & fully                       & two-stage                  & 3D                        &48.5&36.0& 36.5          & 44.9 &42.1       \\
3DVG-Transformer \cite{zhao20213dvg}                                                & ICCV 2021                    & fully                       & two-stage                  & 3D                        &48.5&34.8&34.8& 43.7          & 40.8        \\
FFL-3DOG   \cite{feng2021free}                                                    & ICCV 2021                  & fully                       & two-stage                  & 3D                        &48.2&35.0&37.1& 44.7          & 41.7        \\
InstanceRefer       \cite{yuan2021instancerefer}                                           & ICCV 2021                  & fully                       & two-stage                  & 3D                        &46.0&31.8&34.5& 41.9          & 38.8        \\

3DRefTransformer      \cite{abdelreheem20223dreftransformer}                                         & WACV 2022                  & fully                       & two-stage                  & 3D                        & 46.4              & 32.0            & 34.7     & 41.2&39.0    \\
LanguageRefer     \cite{roh2022languagerefer}                                             & CoRL 2022                  & fully                       & two-stage                  & 3D                        & 51.0              & 36.6            & 41.7     & 45.0&43.9     \\
3D-SPS \cite{luo20223d}                                            & CVPR 2022                & fully                       & one-stage                  & 3D                        & 58.1          & 45.1        & 48.0     & 53.2&51.5     \\
MVT  \cite{huang2022multi}                                                          & CVPR 2022                  & fully                       & two-stage                  & 3D              & 61.3          & 49.1        & 54.3     & 55.4&55.1               \\
BUTD-DETR   \cite{jain2022bottom}                                                   & ECCV 2022                  & fully                       & one-stage                  & 3D                        & -          & -        & -&-     & 49.1    \\

ViL3DRel     \cite{chen2022language}                                                  & NeurIPS 2022               & fully                       & two-stage                  & 3D                       & 70.2          & 57.4        & 62.0     & 64.5&64.4      \\
LAR   \cite{bakr2022look}                                                         & NeurIPS 2022               & fully                       & two-stage                  & 3D                        & 58.4              & 42.3            & 47.4     & 52.1&48.9     \\
3D-VisTA     \cite{zhu20233d}                                                  & ICCV 2023                  & fully                       & two-stage                  & 3D                        &72.1           & 56.7         & 61.5     & 65.1&64.2     \\

EDA    \cite{wu2023eda}                                                  & CVPR 2023                  & fully                       & one-stage                  & 3D                         & -          & -        & -&-     & 52.1     \\
Multi3DRefer      \cite{zhang2023multi3drefer}                                                & CVPR 2023                  & fully                       & two-stage                  & 3D                        & 55.6          & 43.4        & 42.3&52.9     & 49.4     \\

3DRP-Net    \cite{wang20233drp}                                                  & EMNLP 2023                  & fully                       & one-stage                  & 3D                        & 71.4         & 59.7        & 64.2     & 65.2&65.9     \\
CORE-3DVG \cite{yang2024exploiting} $\dagger$                                                     & NeurIPS 2023                  & fully                       & two-stage                  & 3D                        & -          & -        & -&-     & 49.57     \\
G3-LQ \cite{wang2024g}                                                      & CVPR 2024                  & fully                       & one-stage                  & 3D                         & -          & 50.7        & -&-     & 58.4    \\
MA2TransVG \cite{xu2024multi}                                                      & CVPR 2024                  & fully                       & two-stage                  & 3D                         & 71.1          & 57.6        & 62.5&65.4    & 65.2    \\ 
VPP-Net \cite{shi2024aware}                                                       & CVPR 2024                  & fully                       & two-stage                  & 3D                         & -          & -        & -&-     & 56.9\\
MiKASA \cite{chang2024mikasa}                                                       & CVPR 2024                  & fully                       & two-stage                  & 3D                         & 69.7          & 59.4        & 65.4&64.0     & 64.4\\
MCLN \cite{qian2024multi} & ECCV 2024 & fully & one-stage & 3D & - & - & - & - & 59.8 \\
\hline
SAT       \cite{yang2021sat}                                                     & ICCV 2021                  & fully                       & two-stage                  & 3D+2D                     & 56.3          & 42.4        & 46.9&50.4     & 49.2     \\
WS-3DVG      \cite{wang2023distilling}                                                  & ICCV 2023                  & weakly                       & two-stage                  & 3D+2D                     & 27.29          & 17.98        & 21.60    & 22.91&22.45    \\ \hline            
\end{tabular}
\vspace{-10pt}
\label{tab:perform_nr3d}
\end{table*}

\noindent \textbf{Acc@$K$IoU \cite{chen2020scanrefer}.} Intersection over Union (IoU) is a metric commonly used in object detection for measuring similarity between two bounding boxes. Hence, the standard IoU in object detection is defined in a two-dimensional spatial space. However, in the domain of 3D visual grounding, where the goal is to localize objects within a three-dimensional space, a 3D version of IoU is employed. This 3D IoU measures the similarity between ground-truth and predicted bounding boxes in a three-dimensional space. The 3D IoU is calculated as the ratio of the intersection volume to the union volume of the two bounding boxes, with values ranging from 0.0 to 1.0, illustrated in Fig. \ref{fig:figIou}. A higher 3D IoU indicates a better match between the predicted and ground-truth bounding boxes, with an IoU of 1.0 representing an exact match. A leading metric in 3D visual grounding is Acc@$K$IoU \cite{chen2020scanrefer}, which quantifies the proportion of positive predictions that exceed a specified intersection over union (IoU) threshold, denoted as $K$, and typically set at 0.25 or 0.5.

\section{Performance Comparison}
We have summarized the performance of various already-published T-3DVG methods on the ScanRefer, Sr3D, and Nr3D benchmark datasets since 2021 in Tables \ref{tab:perform_scanrefer}, \ref{tab:perform_sr3d}, and \ref{tab:perform_nr3d} in chronological order, respectively. In these tables, we also list the framework type (two-stage or one-stage), the input modality, and the supervision method (fully-supervised or weakly-supervised) for each method.
Overall, two-stage methods tend to perform slightly worse than one-stage methods in grounding due to their high dependency on the performance of pretrained 3D detection models, which can overlook unimportant or inconspicuous objects. However, some recent two-stage methods \cite{yang2024exploiting,chen2022language} have addressed this shortcoming by refining the detected object locations through text guidance or other strategies. For the same method, results are always better when the input includes additional multiview 2D information compared to those using only the 3D modality as input.
In the tables, it can also be observed that methods employing multi-task joint training \cite{cai20223djcg,chen2023unit3d,zhu20233d} often achieve relatively better results. This improvement can be attributed to the benefits of more data and the learning of more comprehensive feature representations brought by learning from different tasks.

By comparing the results across three datasets, it can be observed that the relative performance of methods is generally consistent across different datasets. However, there are subtle differences between the tasks of the ScanRefer dataset and the two datasets of Referit3D (Sr3D, Nr3D). ScanRefer dataset requires methods to regress the bounding box of the target object, while Referit3D datasets only require selecting among the provided bounding boxes. Therefore, some one-stage methods have a more significant advantage on ScanRefer due to their ability to achieve higher quality bounding boxes, whereas this advantage is less noticeable on Sr3D and Nr3D dataset. The comparison between EDA \cite{wu2023eda} and ViL3DRel \cite{chen2022language} provides a clear example of this. Note that the results of the CORE-3DVG \cite{yang2024exploiting} on the Referit3D dataset are based on the detected bounding boxes for grounding, rather than the conventional ground truth bounding boxes, resulting in slightly inferior performance compared to the results on the ScanRefer. 

Additionally, to ensure a fair comparison, we have listed all methods involving large language models (LLMs) in another Table \ref{tab:perform_llm}, regardless of whether LLMs are used for data pre-processing or direct training of multi-modal LLMs. As can be seen from the table, due to the varying use of LLMs, the results of these methods differ significantly. Methods that employ instruction-tuning with multimodal LLMs \cite{yin2024lamm} perform worse compared to common T-3DVG methods; conversely, using LLMs for commonsense reasoning on processed information \cite{fang2023transcribe3d} yields impressive results. This variance in performance underscores the complexity and potential of integrating LLMs in T-3DVG tasks.

\begin{table*}[]
\centering
\caption{Performance Comparison of Methods using LLM.}
\vspace{-10pt}
\begin{tabular}{ccccc|cc|cc}
\hline
\multirow{2}{*}{Method} & \multirow{2}{*}{Venue} & \multirow{2}{*}{Supervise} & \multirow{2}{*}{Category} & \multirow{2}{*}{Modality} & \multicolumn{2}{c|}{ScanRefer} & Sr3D     & Nr3D     \\
                        &                        &                            &                           &                           & Acc@0.25       & Acc@0.5       & Acc & Acc \\ \hline \hline
UniT3D  \cite{chen2023unit3d}                & ICCV 2023              & fully                       & using LLM                 & 3D                        & 45.27          & 39.14         & -        & -        \\
ViewRefer   \cite{guo2023viewrefer}            & ICCV 2023              & fully                       & using LLM                 & 3D                        & 41.35          & 33.69         & 67.2     & 56.1     \\
Transcribe3D   \cite{fang2023transcribe3d}         & CoRL 2023              & fully                       & using LLM                 & 3D                        & -              & -             & 96.6     & 69.4     \\
LAMM   \cite{yin2024lamm}                  & NeurIPS 2024           & fully                       & using LLM                 & 3D                        & -              & 3.38          & -        & -        \\ \hline
\end{tabular}      
\vspace{-10pt}
\label{tab:perform_llm}
\end{table*}

\section{Challenges and Future Directions}
Despite significant performances achieved on benchmark datasets, limitations of T-3DVG methods hinder the practicability and generalization ability of T-3DVG methods. In this section, we first provide the critical analysis of current methods and then discuss potential future directions.
\subsection{Critical Analysis}

\noindent \textbf{Relying on a large amount of annotations.}
Most existing methods for the T-3DVG task are implemented under a fully-supervised setting.
However, in this setting, a huge amount of bounding boxes for numerous objects in the 3D point cloud scenes, along with their descriptive texts, are required to be annotated as labels to provide reliable supervision. 
For example, the widely used ScanRefer dataset \cite{chen2020scanrefer} contains 806 scenes with 51,583 matched object-text labels.
However, 
manually annotating these object-text pairs is very time-consuming and labor-intensive. 
To alleviate this problem, the weakly-supervised methods are proposed to only use text labels for grounding without relying on any object bounding box annotations.
In the weakly-supervised setting, we do not need to annotate bounding boxes for the objects, but descriptive texts still need to be annotated for each object in every scene. 
Although the weakly-supervised methods slightly reduce the annotation cost, tens of thousands of language descriptions still need to be annotated, requiring significant manual effort.
Therefore, how to learn a 3D visual grounding model with as few annotations as possible is essential to investigate.

\noindent \textbf{Complicated background objects and spatial relations.}
Unlike traditional 2D images and videos, 3D scenes contain complicated contextual contents, including multiple classes of objects and their complex spatial relationships.
However, for each sentence input, the 3D grounding models are required to solely output the precise location of a single object accurately. 
Therefore, this raises critical issues for researchers: How to distinguish the ambiguous objects between the background objects and the foreground ones? How to comprehend the spatial relations from the sentence and reason these relationships between the objects with their coordinate contexts in deep networks?
Existing 3D grounding methods simply design specific attention architecture to globally learn the object content and spatial relations from the whole point cloud data, lacking local detail mining for reasoning the special objects in depth.
Some works try to utilize multi-view learning \cite{zhang2023multi3drefer} with contrastive learning paradigms to distinguish objects or spatial relations. However, their methods cost a lot of time and resources to simulate multiple possible views of each 3D point cloud.

\noindent \textbf{Poor practicality.}
3D object grounding endeavors to localize a unique pre-existing object within a single 3D scene given a natural language description. However, 
such a strict setting is unnatural as (1) It is not always possible to know whether a target object actually exists in a specific 3D scene; (2) It cannot comprehend and reason more contextualized descriptions
of multiple objects in more practical 3D cases.
In real-world scenarios, a collection of 3D scenes is generally available, some of which may not contain the described object while some potentially contain multiple target objects. Therefore, existing 3D grounding methods are restricted
to a single-sentence input thus being hard to deploy in practical scenarios.
To this end, it is important to simultaneously process a group of related 3D scenes, allowing a flexible number of target objects to exist in each scene. 

\subsection{Future Directions}

\noindent \textbf{Effective feature extractor.}
Feature quality significantly impacts T-3DVG performance. In the general pipeline, current T-3DVG approaches simply employ separate pre-trained visual and textual extractors to independently extract scene and sentence features. Consequently, there exists a considerable disparity between the extracted visual and textual features residing in distinct feature spaces. Despite efforts by T-3DVG methods to align these features into a unified space, the inherent disparity remains challenging to fully eliminate. Furthermore, disparities can arise between T-3DVG datasets and those used to pre-train feature extractors, potentially resulting in information loss or imprecise representations. 
Considering recent advances in scene-language pre-training \cite{parelli2023clip,zhang2024vision,zhu20233d,jin2023context}, dedicated or more effective feature extractors for T-3DVG are much expected.

\noindent \textbf{Zero-shot T-3DVG.}
As illustrated in the previous subsection, annotations of the T-3DVG task consist of the location (\textit{i.e.}, bounding box) of a specific object in a complicated 3D scene and a corresponding textual sentence. However, obtaining such paired annotation is laborious and expensive in real-world scenarios. 
To avoid the costly annotations, it is essential to propose a zero-shot T-3DVG task setup that aims to learn a 3D grounding model without any paired annotation.
In this setting, the researchers should train a T-3DVG model by leveraging easily accessible unpaired data including 3D scenes, natural language corpora, and an off-the-shelf object detector.
Given an unannotated 3D scene, one should first detect potential 3D objects and develop prompts to generate corresponding pseudo sentences for jointly training the grounding model.
Among them,
how to generate high-quality pseudo sentences by encoding both object classes and spatial correlation is the emergency issue.
Although Yuan \textit{et al.} \cite{yuan2023visual} try to utilize language models to alleviate the lack of text descriptions, it fails to design a well-structured zero-shot learning paradigm for robust representation learning.
Potential solutions can be motivated and adapted from the zero-shot 2D temporal grounding methods \cite{kim2023language,lu2024zero,wang2022prompt,nam2021zero} into the 3D scenarios.

\noindent \textbf{T-3DVG with dense 3D objects.}
The paradigm of previous studies on 3D object grounding is
to simply localize each individual object referred by a single freeform sentence in a 3D scene. Although existing T-3DVG methods have achieved significant progress in recent years, they are limited by the single-sentence input and are not suitable for comprehending and reasoning more contextualized descriptions of multiple objects in complicated 3D scenes. However, such a paragraph of multiple sentences that describe several objects in a specific region of a 3D scene is natural and practical in real-world applications such as robotics and autonomous driving. 
To localize dense objects referred by a paragraph in a 3D scene,
a straightforward idea is to apply well-studied 3D single-object
grounding models to each individual sentence in the paragraph and integrate their results for dense grounding.
However, directly applying them to the multiple-sentence setting
may suffer from two critical issues: (1) Firstly, this solution is naive and only considers a single sentence, failing to learn the contextual semantic relationships among multiple sentences. These contextual semantic relations are vital clues for accurate paragraph comprehension and precise localization of dense objects.  (2) Secondly, this solution fails to leverage the spatial correlations between locations of dense objects described in the
same paragraph for cross-modal spatial reasoning. The existence of
such spatial correlations is due to the fact that humans are likely to
refer linguistically to multiple objects located in a focused region
of a 3D scene rather than describe randomly placed ones. Ignoring
such spatial relations of dense target objects may lead to inferior
performance in precisely finding the location of each one.
Potential solutions can utilize contextual semantic aggregation modules \cite{zeng2020dense,bao2021dense} to jointly comprehend all descriptions and reason different objects among them.

\noindent \textbf{Better integration with large models.}
Multi-modal Large Language Models (MLLMs) have achieved significant success and demonstrated promising capabilities in various 2D multimodal downstream tasks, such as text-to-image generation \cite{nichol2021glide,ramesh2022hierarchical,rombach2022high}, visual question-answering \cite{tsimpoukelli2021multimodal,li2023blip,alayrac2022flamingo}, and \textit{etc.}, due to an increase in the amount of data, computational resources, and number of model parameters. 
By further benefiting from the strong comprehension of large language models (LLMs) \cite{brown2020language,touvron2023llama,touvron2023llama2}, recent 2D MLLMs \cite{dai2024instructblip,liu2024visual,zhu2023minigpt} on top of LLMs show superior performances in solving complex vision-language tasks by utilizing appropriate human-instructed prompts.
By applying the LLMs and pre-trained large 3D vision-language models into the T-3DVG field, it can also provide more representative multimodal features and more accurate cross-modal alignment for complicated scene-based understanding and reasoning.
There are many recently proposed 3D large language models \cite{qi2023gpt4point,tang2024minigpt,hong20233d,fu2024scene}, showing a strong power in handling 3D reasoning tasks.
However, their proposed methods simply follow a general and unified framework that can only address global tasks, such as classification, captioning, generation, and question answering. Their performances are poor on the specific T-3DVG as this task requires detailed local reasoning.
Although some existing T-3DVG works \cite{guo2023viewrefer,yang2023llm,fang2023transcribe3d,jia2024sceneverse} have tried to employ LLMs to individually comprehend and enrich the textual semantics, they still ignore investigating the strong ability of 3D large models to handle multimodal inputs.
Therefore, how to incorporate large models into the specific T-3DVG grounding task with in-depth structure-level designs is an emergency issue.
\section{Conclusion}
Text-guided 3D visual grounding (T-3DVG) is a fundamental yet challenging vision-language task in multimedia understanding.
It is also worth exploring since it serves as an intermediate step for various downstream tasks, such as multi-modal 3D scene
understanding, language-guided robotic grasping, and 3D embodied interaction.
In this survey, we comprehensively review the fundamental elements and recent research advances of T-3DVG.
Specifically, we first provide a general structure of the T-3DVG pipeline with detailed components in a tutorial style. We then summarize the existing T-3DVG approaches into different categories and analyze their strengths and weaknesses. 
Afterwards, we present the benchmark datasets and evaluation metrics to assess their performances. Finally, we discuss the potential limitations of existing T-3DVG and provide some insights in future research.

\bibliographystyle{IEEEtran}
\bibliography{reference.bib}
\end{document}